\documentclass[conference]{IEEEtran}

\RequirePackage[l2tabu, orthodox]{nag} % checks for obsolete LaTeX packages and outdated commands
\usepackage{lipsum}

\usepackage{cite}
\ifCLASSINFOpdf
  \usepackage[pdftex]{graphicx}
  \graphicspath{{./figs/}}
  \DeclareGraphicsExtensions{.pdf,.jpeg,.png}
\else
\fi
\usepackage{adjustbox} %http://tex.stackexchange.com/questions/63229/shift-entire-tikz-picture-vertically-into-part-of-align-environment

\usepackage{tikz}
\usetikzlibrary{decorations.pathreplacing}
\usetikzlibrary{decorations.markings}
\usetikzlibrary{patterns}
\usetikzlibrary{positioning}
\usetikzlibrary{shapes,arrows}
\usetikzlibrary{calc}
\usetikzlibrary{intersections}
\usetikzlibrary{plotmarks}
\usetikzlibrary{spy,backgrounds}
\usepackage{pgfplots}
\usepgflibrary{plotmarks}
\usepgfplotslibrary{units}
\pgfplotsset{compat=1.12}
\usetikzlibrary{pgfplots.groupplots}

\usepackage{mathtools} %extends amsmath and fixes bugs

\usepackage{scalerel,stackengine} % for \equalhat

\usepackage[eulergreek]{sansmath} % sans serif font available in math mode

\usepackage{cleveref}
\crefname{pluralequation}{eqs.}{eqs.}
\Crefname{pluralequation}{Eqs.}{Eqs.}
\crefformat{equation}{#2eq.~(#1)#3}
\Crefformat{equation}{#2Eq.~(#1)#3}
\crefformat{pluralequation}{#2eqs.~(#1)#3}
\Crefformat{pluralequation}{#2Eqs.~(#1)#3}
\crefrangeformat{pluralequation}{eqs.~(#3#1#4) to~(#5#2#6)}
\Crefrangeformat{pluralequation}{Eqs.~(#3#1#4) to~(#5#2#6)}
\crefmultiformat{pluralequation}{eqs.~(#2#1#3)}{ and~(#2#1#3)}{, (#2#1#3)}{ and~(#2#1#3)}
\Crefmultiformat{pluralequation}{Eqs.~(#2#1#3)}{ and~(#2#1#3)}{, (#2#1#3)}{ and~(#2#1#3)}

\ifCLASSOPTIONcompsoc
  \usepackage[caption=false,font=normalsize,labelfont=sf,textfont=sf]{subfig}
\else
  \usepackage[caption=false,font=footnotesize]{subfig}
\fi
\usepackage{booktabs}

\usepackage{wrapfig}

\usepackage{dblfloatfix}    % To enable figures at the bottom of page

\usepackage{placeins} % for \FloatBarrier

\usepackage{capt-of}

\usepackage{siunitx} 
\usepackage[super]{nth} %http://tex.stackexchange.com/questions/4118/whats-the-quickest-way-to-write-2nd-3rd-etc-in-latex

\usepackage{microtype} % improves the spacing between words and letters

\graphicspath{{../figs/}}
\definecolor{Red}{rgb}{1.0,0,0} 
\definecolor{Green}{rgb}{0,.80,0} 
\definecolor{Blue}{rgb}{0,0,0.80}

\newcommand{\classk}{k}
\newcommand{\classK}{K}
\newcommand{\subc}{c}
\newcommand{\subC}{C}
\newcommand{\dimd}{d}
\newcommand{\dimD}{D}
\newcommand{\inpn}{n}
\newcommand{\inpN}{N}
\NewDocumentCommand{\y}{O{}O{}}{y_{#2}^{\,#1}\!}
\NewDocumentCommand{\yVec}{O{}}{\vec{\y}^{\,#1}\!}
\newcommand{\yVecN}{\yVec[(\inpn)]}

\newcommand{\ydN}{\y[\!(\inpn)][\dimd]}
\newcommand{\yd}{\y[][\dimd]\,}
\newcommand{\labely}{l}
\NewDocumentCommand{\uu}{O{}O{}}{u_{#1}^{\,#2}\!}

\NewDocumentCommand{\Wgen}{O{}O{}}{\mathcal{W}_{#1#2}}
\newcommand{\Wgencd}{\Wgen[\subc][\dimd]}
\NewDocumentCommand{\Rgen}{O{}O{}}{\mathcal{R}_{#1#2}}
\newcommand{\Rgenkc}{\Rgen[\classk][\subc]}
\NewDocumentCommand{\W}{O{}O{}}{W_{\!#1#2}}
\newcommand{\Wcd}{\W[\subc][\dimd]}
\newcommand{\WcdN}{\W[\subc][\dimd]^{(\inpn)}}
\NewDocumentCommand{\R}{O{}O{}}{R_{#1#2}}
\newcommand{\Rkc}{\R[\classk][\subc]}

\NewDocumentCommand{\Sc}{O{\subc}O{}}{s_{#1}^{#2}}
\newcommand{\ScN}{\Sc[\subc][(\inpn)]\!}
\NewDocumentCommand{\sVec}{O{}}{\vec{s}_{\vphantom{\subc}}^{\,#1}}

\NewDocumentCommand{\Igenc}{O{\subc}O{}}{\mathcal{I}_{#1}^{#2}}

\NewDocumentCommand{\Ic}{O{\subc}O{}}{I_{#1}^{#2}}
\newcommand{\IcN}{\Ic[\subc][(\inpn)]\!}
\NewDocumentCommand{\tk}{O{\classk}O{}}{t_{#1}^{#2}}

\NewDocumentCommand{\tVec}{O{}}{\vec{t}_{\vphantom{\classk}}^{\,#1}}

\NewDocumentCommand{\eps}{O{}}{\epsilon_{\textnormal{\tiny $#1$}}}
\newcommand{\eW}{\eps[\W]}
\newcommand{\eR}{\eps[\R]}
\NewDocumentCommand{\epst}{O{}}{\tilde{\epsilon}_{\textnormal{\tiny $#1$}}}

\newcommand{\normA}{A}
\newcommand{\BvSB}{\mathrm{BvSB}}

\newcommand{\FF}{\mathcal{F}}
\newcommand{\E}[1]{\big\langle{}#1\big\rangle}

\DeclareMathOperator*{\argmax}{\arg\!\max}

\newcommand\equalhat{\mathrel{\stackon[1.5pt]{=}{\stretchto{%
    \scalerel*[\widthof{=}]{\wedge}{\rule{1ex}{3ex}}}{0.5ex}}}}

\begin{document}
%
% paper title
\title{Truncated Variational EM for \\ Semi-Supervised Neural Simpletrons}
% Truncated Generative Networks for Semi-Supervised Learning

% author names and affiliations
% use a multiple column layout for up to three different
% affiliations
\author{
\IEEEauthorblockN{Dennis Forster}
\IEEEauthorblockA{
  Machine Learning Group\\
  Department for Medical Physics and Acoustics\\
  Carl-von-Ossietzky University of Oldenburg\\
  26129 Oldenburg, Germany\\
  Frankfurt Institute for Advanced Studies (FIAS)\\
  Goethe-University Frankfurt am Main\\
  60438 Frankfurt am Main, Germany\\
  Email: dennis.forster@uol.de}
\and
\IEEEauthorblockN{J\"org L\"ucke}
\IEEEauthorblockA{
  Machine Learning Group\\
  Department for Medical Physics and Acoustics\\
  Cluster of Excellence Hearing4all and\\
  Research Center Neurosensory Sciences\\
  Carl-von-Ossietzky University of Oldenburg\\
  26129 Oldenburg, Germany\\
  Email: joerg.luecke@uol.de}
}

% conference papers do not typically use \thanks and this command
% is locked out in conference mode. If really needed, such as for
% the acknowledgment of grants, issue a \IEEEoverridecommandlockouts
% after \documentclass

% for over three affiliations, or if they all won't fit within the width
% of the page, use this alternative format:
% 
%\author{\IEEEauthorblockN{Michael Shell\IEEEauthorrefmark{1},
%Homer Simpson\IEEEauthorrefmark{2},
%James Kirk\IEEEauthorrefmark{3}, 
%Montgomery Scott\IEEEauthorrefmark{3} and
%Eldon Tyrell\IEEEauthorrefmark{4}}
%\IEEEauthorblockA{\IEEEauthorrefmark{1}School of Electrical and Computer Engineering\\
%Georgia Institute of Technology,
%Atlanta, Georgia 30332--0250\\ Email: see http://www.michaelshell.org/contact.html}
%\IEEEauthorblockA{\IEEEauthorrefmark{2}Twentieth Century Fox, Springfield, USA\\
%Email: homer@thesimpsons.com}
%\IEEEauthorblockA{\IEEEauthorrefmark{3}Starfleet Academy, San Francisco, California 96678-2391\\
%Telephone: (800) 555--1212, Fax: (888) 555--1212}
%\IEEEauthorblockA{\IEEEauthorrefmark{4}Tyrell Inc., 123 Replicant Street, Los Angeles, California 90210--4321}}

% use for special paper notices
\IEEEspecialpapernotice{(Submitted preliminary conference paper. Accepted at IJCNN 2017.)}
%

% make the title area
\maketitle

% As a general rule, do not put math, special symbols or citations
% in the abstract
\begin{abstract}
Inference and learning for probabilistic generative networks is often very challenging and typically prevents scalability to as large networks as used for deep discriminative approaches.
To obtain efficiently trainable, large-scale and well performing generative networks for semi-supervised learning, we here combine two recent developments: a neural network reformulation of hierarchical Poisson mixtures (Neural Simpletrons), and a novel truncated variational EM approach (TV-EM).
TV-EM provides theoretical guarantees for learning in generative networks, and its application to Neural Simpletrons results in particularly compact, yet approximately optimal, modifications of learning equations.
If applied to standard benchmarks, we empirically find, that learning converges in fewer EM iterations, that the complexity per EM iteration is reduced, and that final likelihood values are higher on average. 
For the task of classification on data sets with few labels, learning improvements result in consistently lower error rates if compared to applications without truncation.
Experiments on the MNIST data set herein allow for comparison to standard and state-of-the-art models in the semi-supervised setting.
Further experiments on the NIST SD19 data set show the scalability of the approach when a manifold of additional unlabeled data is available.
\end{abstract}

% no keywords

% For peer review papers, you can put extra information on the cover
% page as needed:
% \ifCLASSOPTIONpeerreview
% \begin{center} \bfseries EDICS Category: 3-BBND \end{center}
% \fi
%
% For peerreview papers, this IEEEtran command inserts a page break and
% creates the second title. It will be ignored for other modes.
\IEEEpeerreviewmaketitle

\section{Introduction}

Truncated approaches have been shown to be a valuable tool for machine learning in a variety of different forms and over a wide application domain \cite{LuckeEggert2010,HennigesEtAl2014,SheikhEtAl2014,DaiLucke2014}:
Truncated posterior approximations are used to overcome infeasible combinatorics and enable competitive performance, e.g., in tasks such as component extraction \cite{HennigesEtAl2014}, source separation and image denoising \cite{SheikhEtAl2014}, or invariant clustering \cite{DaiLucke2014}.
%
% \cite{LuckeEggert2010,SheikhEtAl2014}.
%Heuristic methods like dropout \cite{SrivastavaEtAl2014} or maxout \cite{GoodfellowEtAl2013} have been shown to consistently reduce overfitting effects in deep neural networks and improve classification accuracy in pattern recognition tasks for auditory, visual or textual data.
%Benefits of sparse posteriors to training accuracy and execution speed were recently also observed for Gaussian mixture models \cite{HughesSudderth2016} and it has been proven, that truncated distributions can be %treated within the variational free energy framework \cite{Lucke2016}, allowing for mathematically grounded derivations of new truncated algorithms.
%
In this study, we investigate an application of novel theoretical results for truncated variational distributions \cite{Lucke2016} to Neural Simpletrons~\cite{ForsterEtAl2016}.
These are artificial neural networks that approximate likelihood maximization w.r.t.\ normalized and hierarchical Poisson mixture models.
With an emphasis on unsupervised learning, such networks are especially well suited for settings where only few data points have labels.
Learning from such sparsely labeled data becomes increasingly interesting with the ever increasing amounts of easily available data, while acquisition of data labels generally remains costly and potentially requires a lot of human effort.

In \cref{sec:PoissonMixtures,sec:TV-EM}, we first shortly reiterate the underlying hierarchical generative model and the derived NeSi network.
We then outline the TV-EM algorithm and how to apply it to mixture models in general.
Finally, we use these results to derive a truncated NeSi network and discuss the computational benefits of this approach.

In \cref{sec:NumericalExperiments}, we show numerical experiments on MNIST, where we investigate benefits to convergence times and test errors in semi-supervised settings and compare to state-of-the-art models in this application domain, as well as on the NIST SD19 database to show scalability of the approach.

%From the introduce generative model, we derive truncated variational EM (TV-EM) update steps that are guaranteed to increase the data likelihood.
%Using results from \cite{LuckeSahani2008,KeckEtAl2012,ForsterEtAl2016}, we derive a corresponding generative network that is able to learn semi-supervised in an online fashion via local Hebbian learning rules.
%At convergence, the learned neural weights can be shown to be a close approximation of the converged generative weights with truncated variational EM.
%With the full posterior available as easily computable, closed form solution, we can directly compare the benefits of TV-EM to standard EM learning within the generative networks.
%Numerical experiments on the MNIST data set in semi-supervised settings down to the very limit of a single label per class show significant decreases in test error with truncated learning and a lower dependency on large amounts of labeled data.
%As the model allows for exact calculation of the data likelihood, we can also directly observe both an overall speed-up in likelihood maximization as well as a gain in converged likelihood values for optimal truncated settings.
%Further large scale experiments on the NIST Special Database 19 illustrate the scalability properties of the algorithm.
%
%
%
%
%
%

\section{Learning in Hierarchical Poisson Mixtures}
\label{sec:PoissonMixtures}
%\pagedepth\maxdimen % to prevent page breaks between 'section' and 'wrapfig'
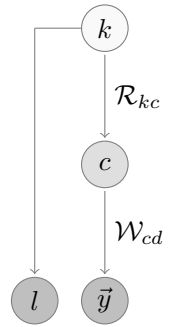
\begin{wrapfigure}[11]{r}{0.145\textwidth}
  \captionsetup{format=plain}
	\centering
	\resizebox{0.12\textwidth}{!}{
		\begin{adjustbox}{trim=0pt 0pt 10pt 8pt}%
			\begin{tikzpicture}[shorten >=1pt,shorten <=1pt,draw=black!50,inner sep=0.333em, outer sep=0.5\pgflinewidth]
	\tikzstyle{neuron}=[circle,fill=black!25,minimum size=17pt,inner sep=0pt]
	\tikzstyle{input neuron}=[neuron, draw = black!50, fill = gray!50]; %gray
	\tikzstyle{subc neuron}=[neuron, draw = black!50, fill = gray!25];%lightgray
	\tikzstyle{classk neuron}=[neuron, draw = black!50, fill = gray!5];%white
	\def\layerdistance{1.75cm}
	\def\nodedistance{0.9cm}
	
	% Draw the input y
	\node[input neuron] (I) at (\nodedistance,0) {$\yVec$};
	
	% Draw the subclass sc
	\node[subc neuron] (C) at (\nodedistance, \layerdistance) {$\subc$};

	% Draw the class tk
	\node[classk neuron] (K) at (\nodedistance, 2*\layerdistance) {$\classk$};
	
	% Draw the label l
	\node[input neuron] (L) at (0,0) {$\labely$};

	% Connect the layers
	\path (K) edge [->] node [right, pos=0.5] {$\Rgenkc$} (C);
	\path (C) edge [->] node [right, pos=0.5] {$\Wgencd$}(I);
	\path (K) edge [-, shorten <= 0pt, shorten >= 0pt] (0, 2*\layerdistance);
	\path (0, 2*\layerdistance) edge [->, shorten <= 0pt] (L);
	%\path (\nodedistance, 1.73*\layerdistance) edge [-, shorten <= 0pt, shorten >= 0pt] (0, 1.73*\layerdistance);
	%\path (0, 1.73*\layerdistance) edge [->, shorten <= 0pt] (L);
\end{tikzpicture}
		\end{adjustbox}
	}
	\caption{Graphical illustration of the hierarchical generative model.}
	\label{fig:generativemodel}
\end{wrapfigure}
\begin{figure}[tbh]
	\centering
	\resizebox{0.92\columnwidth}{!}{
		\begin{adjustbox}{trim=0pt 0pt 0pt 0pt}%
			\begin{tikzpicture}[shorten >=1pt, shorten <= 1pt, draw=black!50, node distance=2.5cm, inner sep=0.333em, outer sep=0.5\pgflinewidth, font=\sffamily]
	\tikzstyle{every pin edge}=[<-,shorten <=1pt]
	\tikzstyle{neuron}=[circle,fill=black!25,color=black!70,thick,minimum size=8pt,inner sep=0pt]
	\tikzstyle{input neuron}=[neuron, draw, fill = gray!50]; %gray
	\tikzstyle{proc1 neuron}=[neuron, draw, fill = gray!25];%lightgray
	\tikzstyle{proc2 neuron}=[neuron, draw, fill = gray!5];%white
	\tikzstyle{annot} = [text width=1em, text centered]
	\def\layerdistance{1.95cm}
	\def\nodedistance{1cm}
	\def\ninput{5}
	\def\nproca{4}
	\def\nprocb{3}
	
	% Draw input
	\node[annot] (I0) at (-\layerdistance*0.75,-\nodedistance/2-\ninput/2*\nodedistance) {$\vec{\tilde{y}}$};

	% Draw the input layer nodes
	\node[annot] (I) at (0,0.5cm-\nodedistance) {$y_d$};
	\foreach \name / \n in {1,...,\ninput}
			\node[input neuron] (I-\name) at (0,-\n*\nodedistance) {};

	% Draw arrow from input to input layer
	\path (I0) edge [->,double,line width=0.6pt,shorten >=8pt,shorten <=4pt] (I-3);
	
	% Draw the first processing layer nodes
	\node[annot] (P1) at (\layerdistance,0.5cm-\ninput/2*\nodedistance + \nproca/2*\nodedistance - \nodedistance) {$s_c$};
	\foreach \name / \n in {1,...,\nproca}
		\node[proc1 neuron] (P1-\name) at (\layerdistance,-\ninput/2*\nodedistance + \nproca/2*\nodedistance - \n*\nodedistance) {};

	% Draw the second processing layer nodes
	\node[annot] (P2) at (2*\layerdistance,0.5cm-\ninput/2*\nodedistance + \nprocb/2*\nodedistance - \nodedistance) {$t_k$};
	\foreach \name / \n in {1,...,\nprocb}
		\node[proc2 neuron] (P2-\name) at (2*\layerdistance,-\ninput/2*\nodedistance + \nprocb/2*\nodedistance - \n*\nodedistance) {};

	% Connect every node in the input layer with every node in the first processing layer.
	\foreach \source in {1,...,\ninput}
		\foreach \dest in {1,...,\nproca}
			\path (I-\source) edge [->, semithick] (P1-\dest);
	\path (I-1) edge node [right, pos=0.3, yshift=0.2cm] {$W_{cd}$} (P1-1);
					
	% Draw lateral interaction of first processing layer
	%\pgfmathparse{int(\nproca-1)}
	%\foreach \source in {1,...,\pgfmathresult}
		%\pgfmathparse{int(\source+1)}
		%\path (P1-\source) edge [<->,line width=0.6pt,dashed] (P1-\pgfmathresult);
	\path (P1-1) edge [<->,line width=0.6pt,dashed] (P1-2);
	\path (P1-2) edge [<->,line width=0.6pt,dashed] (P1-3);
	\path (P1-3) edge [<->,line width=0.6pt,dashed] (P1-4);
	%\path[<->,dash pattern=on 3pt off 2.5pt on 3pt off 2.5pt on 3pt,shorten >=1pt, shorten <=1pt] (P1-1) edge (P1-2);
	%\path[<->,dash pattern=on 3pt off 2.5pt on 3pt off 2.5pt on 3pt,shorten >=1pt, shorten <=1pt] (P1-2) edge (P1-3);
%
	%\path[<->,dash pattern=on 3pt off 2.5pt on 3pt off 2.5pt on 3pt,shorten >=-1.5pt, shorten <=1pt] (P1-\nproca) edge (P1-dots);
	%\path[<->,dash pattern=on 3pt off 2.5pt on 3pt off 2.5pt on 3pt,shorten >=1pt, shorten <=-1.5pt] (P1-dots) edge (P1-n) ;

	% Draw lateral interaction of second processing layer
	\path (P2-1) edge [<->,line width=0.6pt,dashed] (P2-2);
	\path (P2-2) edge [<->,line width=0.6pt,dashed] (P2-3);

	% Connect every node in the first processing layer with every node in the second processing layer.
	\foreach \source in {1,...,\nproca}
		\foreach \dest in {1,...,\nprocb}
			\path (P1-\source) edge [->, semithick] (P2-\dest);
	\path (P1-1) edge node [right, pos=0.3, yshift=0.2cm] {$R_{kc}$} (P2-1);
	%\path[->,semithick,shorten <= 4pt] ($(P1-1.east)+(0,0.05)$) edge ($(P2-1.west)+(0,0.10)$);
	%\path[<-,semithick,shorten >= 4pt] ($(P1-1.east)+(0,0.)$) edge ($(P2-1.west)+(0,0.05)$);
	%\path[->,semithick,shorten <= 4pt] ($(P1-1.east)+(0,0.0)$) edge ($(P2-2.west)+(0,0.10)$);
	%\path[<-,semithick,shorten >= 4pt] ($(P1-1.east)+(0,-0.05)$) edge ($(P2-2.west)+(0,0.05)$);
	%\path[->,semithick,shorten <= 4pt] ($(P1-1.east)+(0,-0.05)$) edge ($(P2-3.west)+(0,0.10)$);
	%\path[<-,semithick,shorten >= 4pt] ($(P1-1.east)+(0,-0.1)$) edge ($(P2-3.west)+(0,0.05)$);
	%%\path[->,semithick] ($(P1-2.east)+(0,0.0)$) edge ($(P2-1.west)+(0,0.1)$);
	%\path[<-,semithick] ($(P1-2.east)+(0,-0.1)$) edge ($(P2-1.west)+(0,0)$);
	%\path[->,semithick] ($(P1-3.east)+(0,0.0)$) edge ($(P2-1.west)+(0,0.1)$);
	%\path[<-,semithick] ($(P1-3.east)+(0,-0.1)$) edge ($(P2-1.west)+(0,0)$);
	%\path[->,semithick] ($(P1-4.east)+(0,0.0)$) edge ($(P2-1.west)+(0,0.1)$);
	%\path[<-,semithick] ($(P1-4.east)+(0,-0.1)$) edge ($(P2-1.west)+(0,0)$);

	% Draw label node
	\node[annot] (L) at (2.75*\layerdistance,-\ninput/2*\nodedistance - 1/2*\nodedistance) {};
	
	%Draw Brackets for Layers
	\draw [decorate,decoration={brace,amplitude=3.5pt, mirror},xshift=0pt,yshift=0]
	(-10pt,-\ninput*\nodedistance-6pt) -- (+10pt,-\ninput*\nodedistance-6pt) node [black,midway,sloped, rotate=0, yshift=-8.5pt] 
	{\scriptsize Obs.};
	\draw [decorate,decoration={brace,amplitude=3.5pt, mirror},xshift=0pt,yshift=0]
	(10pt,-\ninput*\nodedistance-6pt) -- (\layerdistance+10pt,-\ninput*\nodedistance-6pt) node [black,midway,sloped, rotate=0, yshift=-8.5pt] 
	{\scriptsize \nth{1} Hidden};
	\draw [decorate,decoration={brace,amplitude=3.5pt, mirror},xshift=0pt,yshift=0]
	(\layerdistance+10pt,-\ninput*\nodedistance-6pt) -- (2*\layerdistance,-\ninput*\nodedistance-6pt) node [black,midway,sloped, rotate=0, yshift=-8.5pt] 
	{\scriptsize \nth{2} Hidden};
	%\draw [decorate,decoration={brace,amplitude=3.5pt, mirror},xshift=0pt,yshift=8.5pt]
	%(\ninput*\nodedistance+2*\nodedistance + 30.,0) -- (\ninput*\nodedistance+2*\nodedistance + 30.,\layerdistance) node [black,midway,sloped, rotate=180, yshift=8.5pt] 
	%{\scriptsize \nth{1} Hidden};
		%\draw [decorate,decoration={brace,amplitude=3.5pt, mirror},xshift=0pt,yshift=8.5pt]
	%(\ninput*\nodedistance+2*\nodedistance + 30.,\layerdistance) -- (\ninput*\nodedistance+2*\nodedistance + 30.,2*\layerdistance) node [black,midway,sloped, rotate=180, yshift=8.5pt] 
	%{\scriptsize \nth{2} Hidden};
\end{tikzpicture}
		\end{adjustbox}
	}
	\caption{Graphical illustration of the feedforward neural network.}
	\label{fig:neuralnetff}
\end{figure}
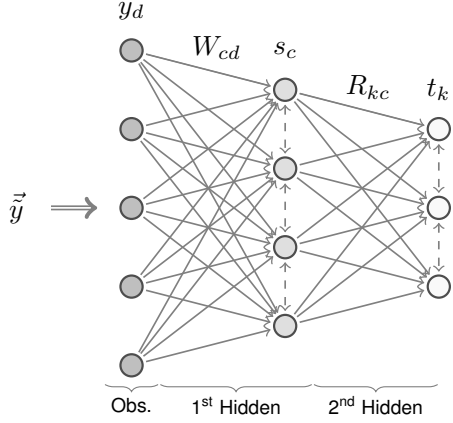

We regard the problem of pattern classification as an inference task in a probabilistic generative model, where we assume the normalized hierarchical Poisson mixture model of \cref{fig:generativemodel} with:
\begin{align}
&\textstyle p(\classk) = \frac{1}{\classK}, \quad\,\, p(\labely|\classk) = \delta_{\labely\classk} 
\label{eq:PriorK}\\[2pt]
&\textstyle p(\subc|\classk,\Rgen) = \Rgenkc%
\label{eq:PriorC}\\[2pt]
&\textstyle p(\yVec\,|\subc,\Wgen) = \prod_\dimd \mathrm{Pois}(\yd;\Wgencd)
\label{eq:Obs}\\[2pt]
&\textstyle \sum_\dimd \Wgencd = \normA,\,\, \sum_\subc \Rgenkc = 1 \,.
\label{eq:WRNormalization}
\end{align}
$(\Wgen,\Rgen)$ herein denote the normalized generative weights, with normalization constants $\normA$ and $1$, respectively.
The labels~$\labely$ are generated directly from the drawn classes~$\classk \in \{1\dots\classK\}$ via a Kronecker-Delta, that is, without any form of label noise.
For the observed data~$\yVec$, first a subclass $\subc \in \{1\dots\subC\}$ belonging to class~$\classk$ is drawn, from which $\yVec$ is generated.
For this generative process, Poisson noise was chosen, which is suitable to model a wide range of non-negative data.
As previously shown~\cite{ForsterEtAl2016}, a hierarchical neural network can be formulated that performs (approximate) maximum-likelihood learning in such a generative model through local Hebbian update rules.
The likelihood objective naturally incorporates learning on labeled and unlabeled data and thus enables semi-supervised learning within a single monolithic algorithm.
The resulting neural network of two hidden layers (see \cref{fig:neuralnetff}), called `Neural Simpletron' (NeSi)~\cite{ForsterEtAl2016}, learns unsupervised subclasses~$\subc$ of pattern representations in the first hidden layer (like different writing styles in case of handwritten character classification), and the classes$~\classk$ of these learned representations in the second hidden layer through provided and self-inferred labels (a.k.a.\ `self-labeling', \cite{ForsterEtAl2016,Lee2013,TrigueroEtAl2015}).

From unnormalized input $\vec{\tilde{\y}}$ to the output probability distribution over classes $\tk$, the activities throughout the network's layers are defined as follows:
\begin{align}
\yd &= \textstyle (\normA-\dimD)\frac{\tilde{\y}_\dimd}{\sum_{\dimd'=1}^{\dimD}\tilde{\y}_{\dimd'}}+1 & \text{normalization, input}
\label{eq:yd}\\
\Ic &= \textstyle \sum_\dimd \log(\Wcd)\yd \span \text{input integration, 1st hidden}
\label{eq:Ic} \\[3pt]
\Sc &= \textstyle \frac{\exp(\Ic\,)}{\sum_{\subc'}\exp(\Ic[\subc'])} & \text{activities, 1st hidden} 
\label{eq:sc}\\
\tk &= \textstyle \left\{\begin{array}{@{\,}lr}
  \delta_{\labely\classk} \span \textnormal{\footnotesize if labeled}\\
  \sum_\subc \! \frac{\Rkc}{\sum_{\classk'} \R[\classk'][\!\subc]} \Sc & \textnormal{\footnotesize else}
\end{array}\right. & \text{activities, 2nd hidden}
\label{eq:tk}
\end{align}
where $(\Wcd)$ are the neural weights from input to 1st hidden layer, and $(\Rkc)$ are the neural weights from 1st to 2nd hidden layer.
During learning, these weights are updated according to the following Hebbian learning rules with implicit subtractive synaptic scaling (see, e.g., \cite{AbbottNelson2000}):
\begin{align}
\Delta\Wcd &= \eW (\Sc \yd - \Sc \Wcd) 
\label{eq:DeltaW}\\
\Delta\Rkc &= \eR (\tk \Sc - \tk \Rkc),
\label{eq:DeltaR}
\end{align}where the top weights $(\Rkc)$ are only updated on labeled data, or on unlabeled data with sufficiently unambiguous inferred labels $\tilde{\labely} = \argmax_k \tk$, that is, if the `Best versus Second Best' \cite{JoshiEtAl2009} lies above a predefined threshold: $\BvSB(\tVec) > \vartheta_{\BvSB}$.

The activities of the neural network herein correspond directly to posterior distributions of the generative model, when taking the neural weights as generative parameters $\Theta = (\W,\R)$:
\begin{align}
\Sc \equalhat p(\subc|\yVec,\Theta),\quad \tk \equalhat \left\{\begin{array}{@{\,}lr}
  p(\classk|\labely\,) \span \text{\footnotesize if labeled}\\
  p(\classk|\yVec,\Theta) & \text{\footnotesize else}
\end{array}\right. .
\end{align}

As shown in \cite{ForsterEtAl2016}, learning neural weights $(\W,\R)$ through \cref{eq:yd,eq:sc,eq:Ic,eq:tk,eq:DeltaW,eq:DeltaR} generally increases the data likelihood under the generative Poisson model and converges to fixed points of the EM algorithm in close approximation (see also \cite{LuckeSahani2008,KeckEtAl2012}).

%----------------- OWN DEFINITIONS -----------------
\newcommand{\qn}{q^{(n)}}
\newcommand{\pn}{p^{(n)}}
\newcommand{\qnPrime}{q^{(n^\prime)}}
\newcommand{\ct}{\tilde{c}}
\newcommand{\qt}{\tilde{q}}
\newcommand{\qnt}{\tilde{q}^{(n)}}
\newcommand{\KK}{\mathcal{K}}
\newcommand{\KKn}{\KK^{(n)}\!}
\newcommand{\KKnew}{\KK^{\mathrm{new}}}
\newcommand{\KKold}{\KK^{\mathrm{old}}}
\newcommand{\KKtilde}{\tilde{\KK}}

\newcommand{\disT}{\textstyle}
\newcommand{\disS}{\displaystyle}

\newcommand{\cNew}{\subc}
\newcommand{\cOld}{\tilde{\subc}}
\newcommand{\cOut}{\tilde{\subc}}
\newcommand{\FFHat}{\hat{\FF}}
\newcommand{\FFHatHat}{\FF}

\newcommand{\IIn}{\mathcal{I}^{(n)}}

%----------------------------------------------------

\section{Truncated Variational EM}
\label{sec:TV-EM}
%\subsection{Truncated Variational Distributions}
Truncated approximations have been introduced to reduce the potentially exponential number of hidden states that have to be evaluated for an exact E-step in generative models with hidden variables.
Truncated approaches are variational EM approximations that do not assume factored variational approximations but are proportional to the exact posteriors in low-dimensional subspaces.
For the purposes of this paper they take the form:
\begin{align}
\qn(\subc;\KK,\Theta)=\frac{p(\subc,\yVecN\,|\Theta)}
{\sum_{\subc'\in\KKn}p(\subc',\yVecN\,|\Theta)}\,\delta(\subc\in\KKn\,),
\label{EqnQMain}
\end{align}
\newdimen\origiwspc%
\origiwspc=\fontdimen2\font% original inter word space
\fontdimen2\font=0.7ex% inter word space
where ${\delta(\subc\in\KKn\,)}$ is an indicator function, i.e., ${\delta(\subc\in\KKn\,)=1}$
\fontdimen2\font=\origiwspc%(original) inter word space
if $\subc\in\KKn$ and zero otherwise.
We have one set of variational parameters $\KKn$ per data point $\yVecN$, and formally define $\KK$ to be the collection of
all these sets: $\KK=(\KK^{(1)},\ldots,\KK^{(\inpN)})$.
The variational distribution gives rise to a free energy of the form
\begin{align}
\FFHatHat(\KK,\Theta) = \sum_{n=1}^{N} \log\!\Big(\smashoperator{\sum_{\hspace{8pt}\subc\in\KKn}} p(\subc,\yVecN\,|\Theta)\Big),
\label{EqnTruncatedF}
\end{align}
which lower-bounds the data log-likelihood w.r.t.\ the generative model defined by $p(\subc,\yVecN\,|\Theta)$ \cite{Lucke2016}.
In general, it is more efficient to optimize the free energy $\FFHatHat(\KK,\Theta)$ instead of the log-likelihood.
For the truncated distributions (\ref{EqnQMain}) the M-step equations remain unchanged compared to the M-steps of the exact posterior, except that expectation values are given by (see \cite{LuckeEggert2010,Lucke2016}):
\begin{equation}
%
%\E{g(\sVec)}_{\qn}\,=\,
\E{g(\subc)}_{\qn(\subc;\KK,\Theta)}=\frac{\sum_{\subc\in\KKn} p(\subc,\yVecN\,|\Theta)\, g(\subc)
}{\sum_{\subc'\in\KKn}p(\subc',\yVecN\,|\Theta)}\,.
\label{EqnSuffStatMain}
\end{equation}
If it can now be shown that the E-step also increases the free energy $\FFHatHat(\KK,\Theta)$ in (\ref{EqnTruncatedF}), then a variational EM algorithm is obtained that increases the lower bound (\ref{EqnTruncatedF}) of the likelihood.

\subsection{Motivation for Mixture Models}
An intuition for applying truncated distributions to mixture models comes from considering common inference results for mixtures.
If we consider, e.g., a Gaussian mixture model (GMM) and typical data distributed according to a GMM, then some clusters will have significant overlap but most cluster components will not overlap substantially.
If the GMM model is fit appropriately to the data, this finally results in very low posterior probabilities for almost all clusters and high posteriors for only very few cluster given any data point.
It seems computationally suboptimal to consider these low posterior probabilities in the same way as the very high ones, since very low probabilities effect inference results very little.
Such numerical operations with small effects can therefore be avoided by setting low posteriors to exactly zero.
Truncated approximations do allow for a formalization of this intuition and, as it will turn out, can provide strong theoretical guarantees and novel algorithmic approaches.

\subsection{TV-EM for Mixture Models}
For mixture models and given a data point $\yVecN$, let us define the set $\KKn$ to consist of $\subC'<\subC$ states. 
In this case, expectation values for the M-step have to be computed based on just these $\subC'$ states in $\KKn$, see \cref{EqnSuffStatMain}.
According to the theoretical results for TV-EM \cite{Lucke2016}, standard M-steps of a mixture model with truncated expectation values (\ref{EqnSuffStatMain}) increase the free energy (\ref{EqnTruncatedF}).
A procedure that also increases the free energy in the E-step is provided by considering that in \cref{EqnTruncatedF} the logarithm is a concave function and its argument is a sum of non-negative probabilities.
Therefore, if we demand that $\subC'$ remains constant, $\FFHatHat(\KK,\Theta)$ increases whenever we replace a state $\cOld\in\KKn$ by a new state $\cNew$ previously not in $\KKn$ such that:
\begin{equation}
p(\cNew,\yVecN\,|\Theta)\,>\,p(\cOld,\yVecN\,|\Theta)\,.
\label{EqnCriterion}
\end{equation}
It is a design choice of the algorithm how much one aims at increasing the free energy based on this criterion. In one extreme, one could terminate updating $\KKn$ in the E-step as soon as one new state is found
that has a larger joint than the lowest joint of the states in $\KKn$. In the other extreme, one could terminate updating $\KKn$
only after the set $\KKn$ is found that maximizes $\FFHatHat(\KK,\Theta)$.
It is typically most promising to use an operation regime in the middle of these two extremes; and for general graphical models it is anyway not possible to find the optimal $\KKn$ efficiently.
For mixture models, however, a full optimization can be obtained because an exhaustive computation of $p(c,\yVecN\,|\Theta)$ for all states (clusters~$\subc$) is possible.
The criterion (\ref{EqnCriterion}) then simply translates to defining $\KKn$ for a given $\yVecN$ such that
\begin{equation}
\forall\subc\!\in\KKn, \forall\cOut\!\not\in\KKn:\, p(\subc,\yVecN\,|\Theta) > p(\cOut,\yVecN\,|\Theta),
\end{equation}
subject to $|\KKn\,|=\subC'$ for all $n$. That is, we define $\KKn$ to consist of the clusters $\subc$ with the $\subC'$ largest joints $p(\cOut,\yVecN\,|\Theta)$.
Such defined sets $\KKn$ necessarily maximize the truncated variational E-step, but require at least a partial sorting, which adds to the computational cost (see below).
Criterion (\ref{EqnCriterion}) on the other hand also allows for more efficient procedures that increase instead of maximize the free energy.
Also the constraint of equally sized $\KKn$ for all $n$ could be relaxed.

\subsection{TV-EM for Neural Simpletrons}
Based on the previous considerations, an application of TV-EM to Neural Simpletrons is straight-forward.
For this, we use the ff-NeSi formulation of \cite{ForsterEtAl2016} which allows for considering the input and first hidden layer to be optimized separately from the second hidden layer (and separately from later self-labeling approaches).
Learning of the weights $\W$ then follows unsupervised likelihood optimization of a (non-hierarchical) normalized Poisson mixture model.
The criterion (\ref{EqnCriterion}) of the Poisson mixture model (observed and first hidden layer) can then be reduced as follows:

\begin{flalign}
&\log\!\Big( \prod_\dimd \mathrm{Pois}(\yd;\Wcd)\,\frac{1}{\subC} \Big) > \log\!\Big( \prod_\dimd \mathrm{Pois}(\yd;\W[\ct][\dimd])\,\frac{1}{\subC} \Big) \nonumber\\[2pt]
&\Leftrightarrow \sum_\dimd  \big( \yd \log(\Wcd) - \log(\yd!) - \Wcd \big) >  \nonumber \\[-10pt] 
&\hspace{100pt}\sum_\dimd  \big( \yd \log( \W[\ct][\dimd] ) - \log(\yd!) - \W[\ct][\dimd] \big) \nonumber\\[2pt]
&\Leftrightarrow \sum_\dimd \! \big( \yd \log(\Wcd) \big) \!-\! \sum_\dimd \! \Wcd > \sum_\dimd \! \big(  \yd \log( \W[\ct][\dimd] ) \big) \!-\! \sum_\dimd \! \W[\ct][\dimd] \nonumber \\[2pt]
&\Leftrightarrow \sum_\dimd \big( \log(\Wcd)\yd \big) = \Ic > \Ic[\ct] = \sum_\dimd \big(\log(\W[\ct][\dimd]) \yd \big)\,,
\end{flalign} 
%
% centered by the '>' sign:
%\begin{align}
%%
%&&\log\!\Big( \prod_\dimd \mathrm{Pois}(\yd;\Wcd)\,\frac{1}{\subC} \Big) &> \log\!\Big( \prod_\dimd \mathrm{Pois}(\yd;\W[\ct][\dimd])\,\frac{1}{\subC} \Big) \nonumber\\[2pt]
%%
%&\Leftrightarrow \sum_\dimd  \big( \yd \log(\Wcd) - \log(\yd!) - \Wcd \big) \span \span \nonumber \\[-12pt] 
%& & &>\nonumber\\[-3pt]
%& & \span \hspace{70pt} \sum_\dimd  \big(  \yd \log( \W[\ct][\dimd] ) - \log(\yd!) - \W[\ct][\dimd] \big)  \nonumber\\[2pt]
%%
%&\Leftrightarrow& \sum_\dimd \! \big( \yd \log(\Wcd) \big) \!-\! \sum_\dimd \! \Wcd &> \sum_\dimd \! \big(  \yd \log( \W[\ct][\dimd] ) \big) \!-\! \sum_\dimd \! \W[\ct][\dimd] \nonumber \\[2pt]
%%
%&\Leftrightarrow& \sum_\dimd \big( \log(\Wcd)\yd \big) = \Ic &> \Ic[\ct] = \sum_\dimd \big(\log(\W[\ct][\dimd]) \yd \big)
%%
%\end{align} 
%
where the last step is a consequence of the normalized weights, \cref{eq:WRNormalization}, used for the mixture model.

It is hence sufficient to only compare the 1st hidden layer inputs $\Ic$ for each data point $\yVecN$ in order to construct sets $\KKn$.
Sets that maximize the free energy in the E-step are consequently obtained simply by selecting those $\subC'$ clusters $\subc$ with the highest values $\Ic$.
Approximate truncated posteriors are then obtained by setting all posteriors $p(\subc|\yVecN,\Theta)$ for $\subc\not\in\KKn$ to zero and renormalizing the non-zero $p(\subc|\yVecN,\Theta)$ to sum to one. 

For the analogy with neural networks, replacing the exact posteriors with truncated posteriors is straight forward.
As the posteriors $p(\subc|\yVecN,\Theta)$ are represented by the activities $\Sc$ of the first hidden layer, the computation of these activities simply changed to take the following form:
\begin{flalign}
\text{compute}\  &\disT \IcN = \sum_\dimd \log(\WcdN)\ydN \quad \text{(as before)}
\label{eq:t_Ic}\\[6pt]
\text{define}\ & \KKn \text{ s.t. } \forall \subc\in\KKn, \forall\cOut\not\in\KKn\,: \IcN > \Ic[\cOut][(\inpn)]
\label{eq:KKn}\\
\text{compute}\  &\ScN =  \frac{\exp(\IcN\,)}{\sum_{\subc'\in\KKn}\exp(\Ic[\subc'][(\inpn)])}\delta(\subc \in \KKn\,),
\label{eq:t_sc}
\end{flalign}
where $\KKn$ is an index set of size $|\KKn\,|=\subC'$.
The results on the equivalence between neural network learning and EM learning directly carry over to truncated learning: now the neural network can be shown to optimize the truncated free energy \cref{EqnTruncatedF} using the truncated inference \cref{eq:t_Ic,eq:KKn,eq:t_sc} and the unchanged learning \cref{eq:DeltaW,eq:DeltaR}.
We will refer to a simpletron with truncated middle layer activations \cref{eq:t_sc} as {\em truncated Neural Simpletron} (t-NeSi). 

\begin{figure*}[!tbh]
    \centering
    \includegraphics[width=\textwidth]{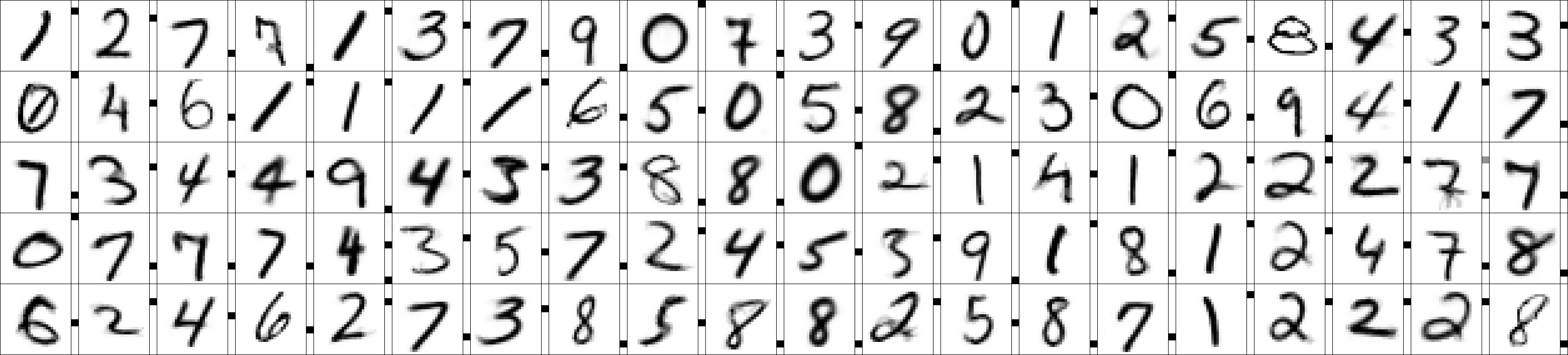}
    \caption{Visualization of a random subset of learned weights by the t-NeSi algorithm trained on MNIST with only a single label per class. Shown are \num{100} of \num{10000} weights $\W[\subc,][:]$ of the first hidden layer as square matrices with weights $\R[:,][\subc]$ of the second hidden layer as columns next to them, indicating the learned class of the individual fields (starting with class `0' at the top of the column and ending with class `9' at the bottom). }
  	\label{fig:MNIST-weights}
\end{figure*}

\subsection{Complexity Reduction per Data Point}
\label{sec:computational_complexity}
Truncated approaches generally reduce the complexity of inference because the number of evaluated hidden states per data point can be drastically reduced (e.g., \cite{LuckeEggert2010,DaiLucke2014,SheikhEtAl2014,Lucke2016}).
For mixture models the reduction of states at first glance does not appear to be very significant (in contrast to multiple causes models) as the number of hidden states scales linearly with the number of hidden variables.
However, the exact zeros for posterior probabilities also result in a large reduction of computational cost in our case.
\Cref{eq:Ic} for $\Ic$ is here still computed fully, which is of $O(\subC\dimD)$.
%But already this operation could be reduced with \cref{EqnCriterion} to \textit{at worst} $O(\subC\dimD)$ and \textit{at best} $O(\subC'\dimD)$ as only values for $\Ic$ have to be found, that increase the free energy.
%
But for the updates of weights $\Wcd$ \cref{eq:DeltaW} the required computations reduce from $O(\subC\dimD)$ to $O(\subC'\dimD)$ after truncation, as those $\Sc$ values that are equal to zero result in no changes for their corresponding weights.
Furthermore, less significantly, the computations of $\Sc$ directly reduces from $O(\subC)$ to $O(\subC')$.

Even with the fully computed $\Ic$, we thus still reduce the computational cost by a number of numerical operations per data point proportional to $(\subC-\subC')\dimD$.
If considering that the additionally needed operations to find the largest $\subC'$ elements are typically of order $O(\subC + \subC' \log \subC)$ per data point \cite{LamTing2000}, we can expect to reduce the required overall operations for t-NeSi by a large fraction compared to non-truncated NeSi networks.

\section{Numerical Experiments}
\label{sec:NumericalExperiments}
For our investigation of t-NeSi, we use, as discussed above, the feedforward NeSi network together with self-labeling, i.e., we refer with t-NeSi to the truncated form of the ff$^+$-NeSi network in \cite{ForsterEtAl2016}.
Using the \textsc{Theano} library \cite{BastienEtAl2012}, the network is implemented for computation on modern GPU hardware.
We show results for the popular MNIST data set, which enables us to also compare to state-of-the-art algorithms in the field of deep learning, as well as the NIST SD19 database as illustration of large scale applicability.

\subsection{MNIST}
The MNIST data set \cite{LeCunEtAl1998} consists of \num{60000} training and \num{10000} test examples of handwritten digits, that have been size-normalized to 28$\times$28 pixel grayscale images and centered by pixel mass.
We investigate semi-supervised settings of 1, 10, 60, 100 and 300 labels per class (with $\classK = 10$~classes total), as well as the fully supervised setting.
Results for t-NeSi are given as mean error rates for the permutation invariant task on the blind test set over 10~independent runs with a new set of randomly picked, class-balanced training labels in each run.
Due to high variance when selecting only a single label per class, 100~runs were performed in this one setting to reduce the error of the mean.

\begin{figure*}[!b]
\setcounter{figure}{4}
\begin{adjustbox}{trim=21pt 2pt 0 0pt}%
	%\tikzsetnextfilename{MNIST-sota}
\pgfplotsset{
  tick label style = {font=\sansmath\sffamily},
  legend style = {font=\sansmath\sffamily},
  label style = {font=\sansmath\sffamily},
	/tikz/font=\sansmath\sffamily,
	discard if not/.style 2 args={
			x filter/.code={
					\edef\tempa{\thisrow{#1}}
					\edef\tempb{#2}
					\ifx\tempa\tempb
					\else
							\def\pgfmathresult{inf}
					\fi
			}
	},
	% #1: index in the group(0,1,2,...)
	% #2: number of plots of that group
	bar group size/.style 2 args={
			/pgf/bar shift={%
							% total width = n*w + (n-1)*skip
							% -> subtract half for centering
							-0.5*(#2*\pgfplotbarwidth + (#2-1)*\pgfkeysvalueof{/pgfplots/bar group skip})  + 
							% the '0.5*w' is for centering
							(.5+#1)*\pgfplotbarwidth + #1*\pgfkeysvalueof{/pgfplots/bar group skip}},%
	},
	bar group skip/.initial=2pt,
	plot 0/.style={color=white,fill=none,mark=none},%
%	plot 1/.style={black!65,fill=black!65,mark=none},%
%	plot 2/.style={black!50,fill=black!50,mark=none},%
%	plot 3/.style={black!25,fill=black!25,mark=none},%
%	plot 4/.style={black!15,fill=black!15,mark=none},%
	plot 1/.style={black,fill=black,mark=none},%
	plot 2/.style={black,pattern color=black,pattern=north east lines,mark=none},%
	plot 3/.style={black!30,fill=black!30,mark=none},%
	plot 4/.style={black!30,pattern color=black!30, pattern=north east lines,mark=none},%
}

\begin{tikzpicture}[scale = 1., font=\sansmath\sffamily]
	\begin{axis}[
		compat=newest, %Better label placement
		width = 1.15/1. * \linewidth, % 1./scale * \linewidth,
		height = 6cm,
		bar width = 0.13cm,
		clip=false,
		ybar,
		ymin = 0,
		ymax = 34,
		ylabel={\fontsize{6}{0}\selectfont\sffamily test error [\%]},
		y label style={at={(0.05,0.42)}},
		y axis line style={draw = none},
		yticklabels={\empty},
		ytick style={draw=none},		
		major grid style=white,
		ymajorgrids,
		xtick=data,
		xtick style={draw=none},
    xticklabels from table={./figs/MNIST-sota-Bar-Label.txt}{Label},		
		xticklabel style={
			inner sep=0pt, 
			anchor=north, 
			rotate=0,
			font = \fontsize{4.8}{0}\selectfont\sffamily,
		},
		every node near coord/.append style={
			color = black!90,
			anchor=mid west, 
			rotate=90,
			font=\fontsize{5}{0}\sansmath\sffamily,
			shift={(axis direction cs:0,0.1)},
			/pgf/number format/precision=2,
			/pgf/number format/fixed,
			/pgf/number format/fixed zerofill,				
    },
		scaled x ticks = false,
		%extra x ticks={3}, extra x tick labels={DBN-rNCA$^\textnormal{\cite{SalakhutdinovHinton2007}}$},
		%extra x ticks={4}, extra x tick labels={PNN},
		%extra tick style={color=red},	
		legend style={
			draw = none,
			at={(0.5,1.2)},
			anchor=north,
			legend columns=0,
			%/tikz/every even row/.append style={row sep=1cm},
			/tikz/every even column/.append style={column sep=0.5cm},
		},
		legend entries={\hspace{1pt}\fontsize{8}{0}\selectfont\sffamily {100} labels,
										\hspace{1pt}\fontsize{8}{0}\selectfont\sffamily {600} labels,
										\hspace{1pt}\fontsize{8}{0}\selectfont\sffamily {1,000} labels,
										\hspace{1pt}\fontsize{8}{0}\selectfont\sffamily {3,000} labels
		},		
		]
		% legend
		\addlegendimage{plot 1}
		\addlegendimage{plot 2}
		\addlegendimage{plot 3}
		\addlegendimage{plot 4}

		% labels
		\addplot [plot 0,bar group size={0}{4}] table [x expr=\coordindex, y=y1] {./figs/MNIST-sota-Bar-Label.txt};
		
		% best results - plots with 4 bars (except DPM)
		\addplot [nodes near coords, plot 1,bar group size={0}{4}, discard if not ={plot}{0}] table [x expr=\coordindex, y=y1] {./figs/MNIST-sota-Bar-Results.txt};
		\addplot [nodes near coords, plot 2,bar group size={1}{4}, discard if not ={plot}{0}] table [x expr=\coordindex, y=y2] {./figs/MNIST-sota-Bar-Results.txt};
		\addplot [nodes near coords, plot 3,bar group size={2}{4}, discard if not ={plot}{0}] table [x expr=\coordindex, y=y3] {./figs/MNIST-sota-Bar-Results.txt};
		\addplot [nodes near coords, plot 4,bar group size={3}{4}, discard if not ={plot}{0}] table [x expr=\coordindex, y=y4] {./figs/MNIST-sota-Bar-Results.txt};
	
		% best results - plots with 2 bars
		\addplot [nodes near coords, plot 2,bar group size={1}{4}, discard if not ={plot}{1}] table [x expr=\coordindex, y=y2] {./figs/MNIST-sota-Bar-Results.txt};
		\addplot [nodes near coords, plot 4,bar group size={3}{4}, discard if not ={plot}{1}] table [x expr=\coordindex, y=y4] {./figs/MNIST-sota-Bar-Results.txt};

		\addplot [nodes near coords, plot 1,bar group size={0}{4}, discard if not ={plot}{2}] table [x expr=\coordindex, y=y1] {./figs/MNIST-sota-Bar-Results.txt};
		\addplot [nodes near coords, plot 3,bar group size={2}{4}, discard if not ={plot}{2}] table [x expr=\coordindex, y=y3] {./figs/MNIST-sota-Bar-Results.txt};

		\addplot [nodes near coords, plot 1,bar group size={0}{4}, discard if not ={plot}{3}] table [x expr=\coordindex, y=y1] {./figs/MNIST-sota-Bar-Results.txt};

		\draw[very thick, color=black!40] (axis cs:-0.4,-7) -- (axis cs:4.4,-7);
		\node[] at (2,-10) {\fontsize{6}{0}\selectfont\sffamily {+10,000} labels};
		\draw[very thick, color=black!40] (axis cs:4.6,-7) -- (axis cs:11.4,-7);
		\node[] at (8,-10) {\fontsize{6}{0}\selectfont\sffamily {+1,000} labels};
	\end{axis}
\end{tikzpicture}
\end{adjustbox}
%\begin{adjustbox}{trim=20pt 0 0 0}%
%	\input{./figs/MNIST-complex-Bar}
%\end{adjustbox}
\caption{Comparison of different algorithms on MNIST data with few labels.
The figure shows results for systems using 100, 600, 1000, and 3000 labeled data points for training.
All algorithms except ours use \num{1000} or \num{10000}~additional data labels (from the training or test set) for parameter tuning.
%The bottom figure gives the number of tunable parameters (as estimated in \cref{TabHyperParams}) and, where known, learned parameters of the algorithms (note the different scales).
%The algorithms also use different tuning, training and testing  protocols.
%\Cref{sec:ProtocolComparison} gives more details. %appendix~D
%\vspace{-2mm}
}
\label{fig:MNIST-sota-Bar}
\vspace{8pt}
\end{figure*}

\subsubsection{Parameter Tuning}
For better direct comparability, we take the same parameter setting for t-NeSi as for its non-truncated counterpart ff$^+$-NeSi in \cite{ForsterEtAl2016}, which was optimized using 10~labeled training examples per class, with a half/half split of labeled data into training and validation set.
We here only optimize the new free parameter $\subC'$, using the same data split.
The resulting optimized free parameters are given as: $\normA=900$, $\subC=\num{10000}$, $\subC'=15$, $\eW=0.2 \times \subC/\inpN$, $\eR=0.2 \times \classK/\inpN$, $\vartheta_{\BvSB}=0.6$, with $\inpN$ being the number of available training samples and $\classK=10$ classes.
To allow for sufficient weight convergence, 500~training iterations over the complete given training sets were performed.
For the single-label/class case this number was increased to 2000~training iterations, as with so few labels more iterations of the top layer were necessary until convergence.
An example of learned weights is shown in \cref{fig:MNIST-weights}.

\subsubsection{Convergence}
Complementary to the reduced computational complexity (see \cref{sec:computational_complexity}), we also observe significantly faster learning times of truncated networks.
\Cref{fig:MNIST-likelihood} shows the training likelihood of only the first hidden layer of truncated and non-truncated NeSi for 10~independent runs each (which are however hardly distinguishable from another at this scale, as the likelihoods within each setting are too close together).
As can be seen, the likelihood increases faster with lower $\subC'$.
However, when the posterior is truncated too much, the likelihood will converge to significantly lower values.
Notably, we can here also observe that the optimal setting $\subC'=15$, found via optimization on a validation set, achieves a higher likelihood than all other shown settings, which could possibly allow for parameter tuning based solely on the (unsupervised) likelihood.

\noindent\begin{minipage}{\columnwidth}
\setcounter{figure}{3}
%\begin{figure}[!hbt]
    \vspace*{\textfloatsep}% Insert "regular" gap between float and text
    \centering
    \begin{adjustbox}{trim=6pt 6pt 0pt 4pt}%
      \resizebox{0.95\linewidth}{!} {
      		\includegraphics{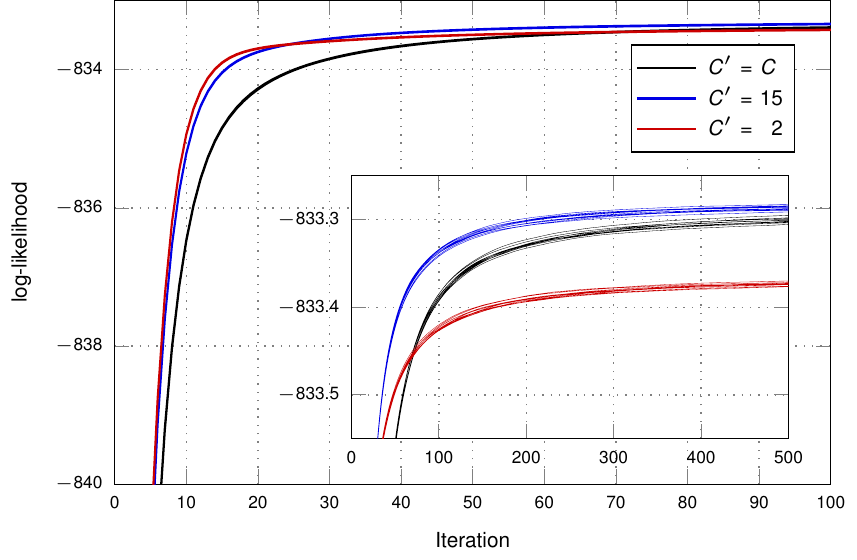}
      	}
    \end{adjustbox}
    \captionof{figure}{Likelihood of first hidden layer for different degrees of truncation. Note the different scalings of the x- and y-axes in the inlaid plot.}
    %\vspace*{\textfloatsep}% Insert "regular" gap between float and text
  	\label{fig:MNIST-likelihood}
%\end{figure}
\setcounter{figure}{5}
\end{minipage}

\subsubsection{Results}

In addition to reducing the number of required numerical operations, we observe that the performance of t-NeSi does not decrease.
Instead, we even observe a significant increase of performance which is reflected on MNIST by significant decreases of error rates in all investigated settings.
\Cref{tab:MNIST-results} shows the results and comparison with non-truncated NeSi.
Although the truncation approach is here only used in a feedforward network with a completely unsupervised middle layer, which thus is independent of the amount of available labels, we still see strong improvements especially in the settings of very few labels.
This shows, that learning of clear and distinct clusters in the unsupervised middle layer becomes particularly crucial for reliable class-learning in the semi-supervised top layer when labeled data is very sparse.

The main source of high error rates in these settings of very limited amounts of labels (down to a single label per class) stems from labeled training data that lie very much on the boundaries of the learned clusters, which can lead to class confusion (e.g.\ all `5's are learned as `8's and vice versa).
With better defined clusters, this confusion appears less frequently, resulting in overall decreased mean error rates.

\begin{table}[bth]
  \centering
  %\vspace{4pt}
  {
\fontsize{8}{10} \sffamily \sansmath
\renewcommand{\arraystretch}{1.3}
\newcommand{\midspace}{\hphantom{abc}}
\setlength{\tabcolsep}{12pt}
\scriptsize
\begin{tabular}{@{}l r@{}c@{}l r@{}c@{}l}
%\bottomrule
\toprule
\#labels (total) & \multicolumn{3}{c}{ff\textsuperscript{+}-NeSi \cite{ForsterEtAl2016}} & \multicolumn{3}{c}{t-NeSi}\\
\midrule
10   & 10.91 &$\,\pm\,$& 0.86 (8.64) & \bfseries 7.22 &$\,\pm\,$& 0.53 (5.33)\\
100  & 4.96 &$\,\pm\,$& 0.08 (0.82) & \bfseries 3.93 &$\,\pm\,$& 0.08 (0.24)\\
600  & 4.08 &$\,\pm\,$& 0.02 (0.17) & \bfseries 3.69 &$\,\pm\,$& 0.03 (0.11)\\
1000 & 4.00 &$\,\pm\,$& 0.01 (0.12) & \bfseries 3.71 &$\,\pm\,$& 0.02 (0.06)\\
3000 & 3.85 &$\,\pm\,$& 0.01 (0.11) & \bfseries 3.56 &$\,\pm\,$& 0.03 (0.11)\\
60000& 3.27 &$\,\pm\,$& 0.01 (0.08) & \bfseries 2.93 &$\,\pm\,$& 0.03 (0.08)\\
\bottomrule
\end{tabular}
\setlength{\tabcolsep}{6pt}
}

  \vspace{10pt}  
  \caption{
    Results on permutation invariant MNIST for different semi-supervised settings using the feedforward Neural Simpletron with and without truncation.
    Given are the mean test error as well as the standard error of the mean (SEM) with the standard deviation (STD) in parentheses.
  }
  \vspace{-10pt}  
  \label{tab:MNIST-results}
\end{table}

\subsubsection{Comparison to State-of-the-Art}
\begin{figure*}[!t]
  \centering
	\begin{adjustbox}{trim=20pt 0pt 0 0}%
   	%\tikzsetnextfilename{MNIST-sota-Scatter1}
\pgfplotsset{
	grid style={dotted,gray},
	minor grid style={dotted,lightgray},
  tick label style = {font=\tiny\sansmath\sffamily},
  legend style = {font=\sansmath\sffamily},
  xlabel style = {font=\sansmath\sffamily},
  ylabel style = {font=\sansmath\sffamily},
	% to match the colors of the markers to the plot cycle list 'exotic'
}
\pgfplotscreateplotcyclelist
{mark list}{%
every mark/.append style={fill=.!80!black},
mark=*\\%
every mark/.append style={fill=.!80!black},
mark=square*\\%
every mark/.append style={fill=.!80!black},
mark=triangle*\\%
mark=star\\%
every mark/.append style={fill=.!80!black},
mark=diamond*\\%
every mark/.append style={fill=.!80!black},
mark=otimes*\\%
mark=+\\%
every mark/.append style={fill=.!80!black},
mark=pentagon*\\%
}
\begin{tikzpicture}
\tikzstyle{annot}=[font=\fontsize{4}{0}\selectfont\sffamily]; %gray
\tikzset{mark size={1.0}}
\begin{groupplot}[
  group style={
    group name=my fancy plots,
    group size=1 by 2,
    %yticklabels at=edge left,
    %xticklabels at=edge bottom,
    vertical sep=0pt,
    horizontal sep=0pt,
  },
	xmin=0, xmax=13250,
  %domain=120:13000,
	clip=false,
	cycle list name=mark list,
	%x dir=reverse,	
  x=0.0015cm,      % set x scale (for width)
  axis on top
]

\nextgroupplot[
  height=2.2cm,
	%yminorgrids,
	%ymajorgrids,
  ymin=-1,ymax=0,
  ytick={0},
  %y dir=reverse,	
  y axis line style=-stealth,
  axis y discontinuity=parallel,
  axis y line=left,
  xmin=0,xmax=3250,
  axis x line=none,
  %xtick={60,80},
  %x axis line style=-, % switch off the axis arrow tips,
	scaled x ticks = false,
  %width=2.0cm, % don't set width,
  %x=0.1cm,      % set x scale (for width)
]
\draw[dashed,red,thick] (axis cs:18,-13) -- (axis cs:18,0); 
\node[circle,fill,outer sep=0pt,red,scale=0.4] at (0,0) (1){};
\node[align=left, font=\fontsize{4}{0}\selectfont\sffamily] at (525,-0.5) (2) {optimal\\[0pt]limit};
\draw[shorten <=3pt,<-, red, thick] (0,0) -- (230,-0.29);
\draw[dashed,red,thick] (axis cs:0,0) -- (axis cs:\pgfkeysvalueof{/pgfplots/xmax},0); 
\nextgroupplot[
  height=9cm,
	yminorgrids,
	ymajorgrids,
	ymode=log,
	ytick={0.5,1,2,5,10,20},
	minor ytick={0.5,0.6,0.7,0.8,0.9,1,2,3,4,5,6,7,8,9,10,20},
	scaled y ticks = false,
	log ticks with fixed point,
  ymin=0.75, ymax=30,
	ylabel={\fontsize{6}{0}\selectfont\sffamily test error [\%]},
	ylabel near ticks, yticklabel pos=left,
  y dir=reverse,	
  xmin=0,xmax=3250,
  axis y line=left,
  y axis line style=-,
  %axis x discontinuity=parallel, % disc. is at start, so avoid for first
  axis x line=bottom,
	xlabel={\fontsize{6}{0}\selectfont\sffamily \#labels for training},
	%xlabel near ticks, xticklabel pos=lower,
	x label style={at={(0.5,-0.09)}},		
  %width=4.5cm, % don't set width,
]
%\addplot {0*x};
\addplot+ [color=blue!50!white,mark=o] table[x=Label, y=ff+-NeSi] {./figs/MNIST-sota-Scatter1-Results.txt};
\addplot+ [color=blue!80!black,mark=*] table[x=Label, y=t-ff+-NeSi] {./figs/MNIST-sota-Scatter1-Results.txt};
\addplot+ [red!70!black,mark=triangle*] table[x=Label, y=VAT] {./figs/MNIST-sota-Scatter1-Results.txt};
\addplot+ [brown!80!black,mark=|] table[x=Label, y=Ladder] {./figs/MNIST-sota-Scatter1-Results.txt};
\addplot+ [color=orange!80!black,mark=asterisk] table[x=Label, y=EmbedCNN] {./figs/MNIST-sota-Scatter1-Results.txt};
\addplot+ [color=green!50!black,mark=diamond*] table[x=Label, y=TSVM] {./figs/MNIST-sota-Scatter1-Results.txt};
\addplot+ [color=lime!70!black,mark=diamond] table[x=Label, y=SVM] {./figs/MNIST-sota-Scatter1-Results.txt};
\addplot+ [color=red!70!black!60!white,mark=+] table[x=Label, y=NN] {./figs/MNIST-sota-Scatter1-Results.txt};
\addplot+ [darkgray,mark=square*] table[x=Label, y=M1+M2] {./figs/MNIST-sota-Scatter1-Results.txt};
\addplot+ [magenta!80!black,mark=star] table[x=Label, y=AtlasRBF] {./figs/MNIST-sota-Scatter1-Results.txt};
\addplot+ [yellow!60!orange!80!black,mark=Mercedes star] table[x=Label, y=AGR] {./figs/MNIST-sota-Scatter1-Results.txt};
\addplot+ [red!50!black!80!white,mark=pentagon*] table[x=Label, y=DBN-rNCA] {./figs/MNIST-sota-Scatter1-Results.txt};
\node[annot,anchor=west,xshift=-1pt,color=orange!80!black] at (3000,1.83) {EmbedCNN};
\node[annot,anchor=west,xshift=-1pt,yshift=0pt,color=red!70!black] at (3000,1.24) {VAT};
\node[annot,anchor=west,xshift=-1pt,yshift=+0pt,color=brown!80!black] at (1000,0.84) {Ladder};
\node[annot,anchor=west,xshift=-1pt,color=darkgray] at (3000,2.18) {M1+M2};
\node[annot,anchor=west,xshift=-1pt,yshift=4pt,color=red!50!black!80!white] at (3000,3.30) {DBN-rNCA};
\node[annot,anchor=west,xshift=-1pt,yshift=2.25pt,color=green!50!black] at (3000,3.45) {TSVM};
\node[annot,anchor=west,xshift=-1pt,yshift=0.4pt,color=blue!80!black] at (3000,3.56) {t-NeSi};
\node[annot,anchor=west,xshift=-1pt,yshift=0pt,color=blue!50!white] at (3000,3.85) {ff\textsuperscript{+}-NeSi};
\node[annot,anchor=west,xshift=-1pt,yshift=-0.25pt,color=lime!70!black] at (3000,4.21) {SVM};
\node[annot,anchor=west,xshift=-1pt,color=red!70!black!60!white] at (3000,6.04) {NN};
\node[annot,anchor=west,xshift=-1pt,yshift=3pt,color=magenta!80!black] at (1000,3.68) {AtlasRBF};
\node[annot,anchor=west,xshift=-1pt,yshift=0pt,color=yellow!60!orange!80!black] at (1000,6.17) {AGR};
\end{groupplot}
\end{tikzpicture}
	\end{adjustbox}
	\begin{adjustbox}{trim=0 0pt 0 0}%
  	  %\tikzsetnextfilename{MNIST-sota-Scatter2}
\pgfplotsset{
	grid style={dotted,gray},
	minor grid style={dotted,lightgray},
  tick label style = {font=\tiny\sansmath\sffamily},
  legend style = {font=\sansmath\sffamily},
  xlabel style = {font=\sansmath\sffamily},
  ylabel style = {font=\sansmath\sffamily},
	% to match the colors of the markers to the plot cycle list 'exotic'
}
\pgfplotscreateplotcyclelist
{mark list}{%
every mark/.append style={fill=.!80!black},
mark=*\\%
every mark/.append style={fill=.!80!black},
mark=square*\\%
every mark/.append style={fill=.!80!black},
mark=triangle*\\%
mark=star\\%
every mark/.append style={fill=.!80!black},
mark=diamond*\\%
every mark/.append style={fill=.!80!black},
mark=otimes*\\%
mark=+\\%
every mark/.append style={fill=.!80!black},
mark=pentagon*\\%
}
\begin{tikzpicture}
\tikzstyle{annot}=[font=\fontsize{4}{0}\selectfont\sffamily]; %gray
\tikzset{mark size={1.0}}
\begin{groupplot}[
  group style={
    group name=my fancy plots,
    group size=2 by 2,
    %yticklabels at=edge left,
    %xticklabels at=edge bottom,
    vertical sep=0pt,
    horizontal sep=0pt,
  },
	xmin=0, xmax=13250,
  %domain=120:13000,
	clip=false,
	cycle list name=mark list,
	%x dir=reverse,	
  x=0.0011cm,      % set x scale (for width)
  axis on top
]

\nextgroupplot[
  height=2.2cm,
	%yminorgrids,
	%ymajorgrids,
  ymin=-1,ymax=0,
  ytick={0},
  %y dir=reverse,	
  y axis line style=-stealth,
  axis y discontinuity=parallel,
  axis y line=left,
  xmin=0,xmax=4350,
  axis x line=none,
  %xtick={60,80},
  %x axis line style=-, % switch off the axis arrow tips,
	scaled x ticks = false,
  %width=2.0cm, % don't set width,
  %x=0.1cm,      % set x scale (for width)
]
\draw[dashed,red,thick] (axis cs:25,-13) -- (axis cs:25,0) {};
\node[circle,fill,outer sep=0pt,red,scale=0.4] at (0,0) (1){};
\node[align=left, font=\fontsize{4}{0}\selectfont\sffamily] at (800,-0.5) (2) {optimal\\[0pt]limit};
\draw[shorten <=3pt,<-, red, thick] (0,0) -- (370,-0.3);
\nextgroupplot[
  height=2.2cm,
  ymin=-1,ymax=0,
  axis y line=none,
  xmin=9100,xmax=13500,
  axis x line=none,
  %xtick={60,80},
  %x axis line style=-, % switch off the axis arrow tips,
	scaled x ticks = false,
  %width=2.0cm, % don't set width,
  %x=0.1cm,      % set x scale (for width)
]
\draw[dashed,red,thick] (axis cs:4850,0) -- (axis cs:\pgfkeysvalueof{/pgfplots/xmax},0); 
\nextgroupplot[
  height=9cm,
	ymode=log,
	ytick={0.5,1,2,5,10,20},
	minor ytick={0.5,0.6,0.7,0.8,0.9,1,2,3,4,5,6,7,8,9,10,20},
	scaled y ticks = false,
	log ticks with fixed point,
  ymin=0.75, ymax=30,
%	ylabel={\fontsize{6}{0}\selectfont\sffamily test error [\%]},
%	ylabel near ticks, yticklabel pos=left,
  y dir=reverse,	
  xmin=0,xmax=4350,
  axis y line=left,
  y axis line style=-,
  %axis x discontinuity=parallel, % disc. is at start, so avoid for first
  axis x line=bottom,
  x axis line style=-,  
  %x axis line style=stealth-, % let arrow head point to left	
	xlabel={\fontsize{6}{0}\selectfont\sffamily total \#labels for training and tuning},
	%xlabel near ticks, xticklabel pos=lower,
	x label style={at={(1.0,-0.09)}},		
  %width=4.5cm, % don't set width,
]
%\addplot {0*x};
\addplot+ [color=blue!80!black,mark=*] table[x=Label, y=t-ff+-NeSi] {./figs/MNIST-sota-Scatter2-Results.txt};
\addplot+ [color=blue!50!white,mark=o] table[x=Label, y=ff+-NeSi] {./figs/MNIST-sota-Scatter2-Results.txt};
\addplot+ [color=red!70!black,mark=triangle*] table[x=Label, y=VAT] {./figs/MNIST-sota-Scatter2-Results.txt};
\addplot+ [color=orange!80!black,mark=asterisk] table[x=Label, y=EmbedCNN] {./figs/MNIST-sota-Scatter2-Results.txt};
\addplot+ [color=green!50!black,mark=diamond*] table[x=Label, y=TSVM] {./figs/MNIST-sota-Scatter2-Results.txt};
\addplot+ [color=lime!70!black,mark=diamond] table[x=Label, y=SVM] {./figs/MNIST-sota-Scatter2-Results.txt};
\addplot+ [color=red!70!black!60!white,mark=+] table[x=Label, y=NN] {./figs/MNIST-sota-Scatter2-Results.txt};
\node[annot,anchor=north west,yshift=2pt,xshift=-4pt,color=blue!50!white] at (100,4.96) {ff\textsuperscript{+}-NeSi};
\node[annot,anchor=north west,yshift=10.5pt,xshift=-4pt,color=blue!80!black] at (100,3.93) {t-NeSi};
\node[annot,anchor=north west,yshift=1.5pt,xshift=-8pt,color=red!70!black] at (1100,2.12) {VAT};
\node[annot,anchor=north,yshift=1.5pt,xshift=-2pt,color=orange!80!black] at (1100,7.75) {EmbedCNN};
\node[annot,anchor=north east,yshift=1.5pt,xshift=5pt,color=green!50!black] at (1100,16.81) {TSVM};
\node[annot,anchor=east,xshift=2pt,color=lime!70!black] at (1100,23.44) {SVM};
\node[annot,anchor=west,xshift=-2pt,color=red!70!black!60!white] at (1100,25.81) {NN};
% draw artificial grid
\draw[dotted,gray] (axis cs:0,1) -- (axis cs:\pgfkeysvalueof{/pgfplots/xmax},1); 
\draw[dotted,gray] (axis cs:0,2) -- (axis cs:\pgfkeysvalueof{/pgfplots/xmax},2); 
\draw[dotted,gray] (axis cs:0,5) -- (axis cs:\pgfkeysvalueof{/pgfplots/xmax},5); 
\draw[dotted,gray] (axis cs:0,10) -- (axis cs:\pgfkeysvalueof{/pgfplots/xmax},10); 
\draw[dotted,gray] (axis cs:0,20) -- (axis cs:\pgfkeysvalueof{/pgfplots/xmax},20); 
\draw[dotted,lightgray] (axis cs:0,0.8) -- (axis cs:\pgfkeysvalueof{/pgfplots/xmax},0.8); 
\draw[dotted,lightgray] (axis cs:0,0.9) -- (axis cs:\pgfkeysvalueof{/pgfplots/xmax},0.9); 
\draw[dotted,lightgray] (axis cs:0,3) -- (axis cs:\pgfkeysvalueof{/pgfplots/xmax},3); 
\draw[dotted,lightgray] (axis cs:0,4) -- (axis cs:\pgfkeysvalueof{/pgfplots/xmax},4); 
\draw[dotted,lightgray] (axis cs:0,6) -- (axis cs:\pgfkeysvalueof{/pgfplots/xmax},6); 
\draw[dotted,lightgray] (axis cs:0,7) -- (axis cs:\pgfkeysvalueof{/pgfplots/xmax},7); 
\draw[dotted,lightgray] (axis cs:0,8) -- (axis cs:\pgfkeysvalueof{/pgfplots/xmax},8); 
\draw[dotted,lightgray] (axis cs:0,9) -- (axis cs:\pgfkeysvalueof{/pgfplots/xmax},9); 
\nextgroupplot[
  height=9cm,
	ymode=log,
	ytick={0.5,1,2,5,10,20},
	minor ytick={0.5,0.6,0.7,0.8,0.9,1,2,3,4,5,6,7,8,9,10,20},
	scaled y ticks = false,
	log ticks with fixed point,
  ymin=0.75, ymax=30,
  y dir=reverse,	
  xmin=9100,xmax=13500,
  axis y line=none,
  axis x discontinuity=parallel,
  axis x line=bottom,
  %x axis line style=-, % switch off the axis arrow tips,
	scaled x ticks = false,
  %width=2.0cm, % don't set width,
  %x=0.1cm,      % set x scale (for width)
]
\addplot+ [brown!80!black,mark=|] table[x=Label, y=Ladder] {./figs/MNIST-sota-Scatter2-Results.txt};
\addplot+ [darkgray, mark=square*] table[x=Label, y=M1+M2] {./figs/MNIST-sota-Scatter2-Results.txt};
\addplot+ [magenta!80!black, mark=star] table[x=Label, y=AtlasRBF] {./figs/MNIST-sota-Scatter2-Results.txt};
\addplot+ [yellow!60!orange!80!black, mark=Mercedes star] table[x=Label, y=AGR] {./figs/MNIST-sota-Scatter2-Results.txt};
\addplot+ [red!50!black!80!white,mark=pentagon*] table[x=Label, y=DBN-rNCA] {./figs/MNIST-sota-Scatter2-Results.txt};
\node[annot,anchor=north, yshift=-12pt,color=brown!80!black] at (10100,0.84) {Ladder};
\node[annot,anchor=north, yshift=1.5pt,color=darkgray] at (10100,3.33) {M1+M2};
\node[annot,anchor=east,xshift=1pt,color=magenta!80!black] at (10100,8.10) {AtlasRBF};
\node[annot,anchor=west,xshift=-8.5pt,yshift=-4pt,color=red!50!black!80!white] at (10600,8.70) {DBN-rNCA};
\node[annot,anchor=east,xshift=1pt,color=yellow!60!orange!80!black] at (10100,9.40) {AGR};
% draw artificial grid
\draw[dotted,gray] (axis cs:9100,1) -- (axis cs:\pgfkeysvalueof{/pgfplots/xmax},1); 
\draw[dotted,gray] (axis cs:9100,2) -- (axis cs:\pgfkeysvalueof{/pgfplots/xmax},2); 
\draw[dotted,gray] (axis cs:9100,5) -- (axis cs:\pgfkeysvalueof{/pgfplots/xmax},5); 
\draw[dotted,gray] (axis cs:9100,10) -- (axis cs:\pgfkeysvalueof{/pgfplots/xmax},10); 
\draw[dotted,gray] (axis cs:9100,20) -- (axis cs:\pgfkeysvalueof{/pgfplots/xmax},20); 
\draw[dotted,lightgray] (axis cs:9100,0.8) -- (axis cs:\pgfkeysvalueof{/pgfplots/xmax},0.8); 
\draw[dotted,lightgray] (axis cs:9100,0.9) -- (axis cs:\pgfkeysvalueof{/pgfplots/xmax},0.9); 
\draw[dotted,lightgray] (axis cs:9100,3) -- (axis cs:\pgfkeysvalueof{/pgfplots/xmax},3); 
\draw[dotted,lightgray] (axis cs:9100,4) -- (axis cs:\pgfkeysvalueof{/pgfplots/xmax},4); 
\draw[dotted,lightgray] (axis cs:9100,6) -- (axis cs:\pgfkeysvalueof{/pgfplots/xmax},6); 
\draw[dotted,lightgray] (axis cs:9100,7) -- (axis cs:\pgfkeysvalueof{/pgfplots/xmax},7); 
\draw[dotted,lightgray] (axis cs:9100,8) -- (axis cs:\pgfkeysvalueof{/pgfplots/xmax},8); 
\draw[dotted,lightgray] (axis cs:9100,9) -- (axis cs:\pgfkeysvalueof{/pgfplots/xmax},9); 
\end{groupplot}
\end{tikzpicture}
	\end{adjustbox}
\caption{
Classification performance of different algorithms compared against varying proportion of labeled training data.
The left-hand-side plot shows the achieved test errors w.r.t.\ the amount of labeled data seen by the compared algorithms during training.
The right-hand-side plot illustrates for the same experiments the total amount of labeled data seen by each of the algorithms over the whole tuning and training procedure.
The plots can be read similar to ROC curves, in the way that the more a curve approaches the upper-left corner, the better is the performance of a system for decreasing amounts of available labeled data.
}
\label{fig:MNIST-sota-Scatter}
\end{figure*}
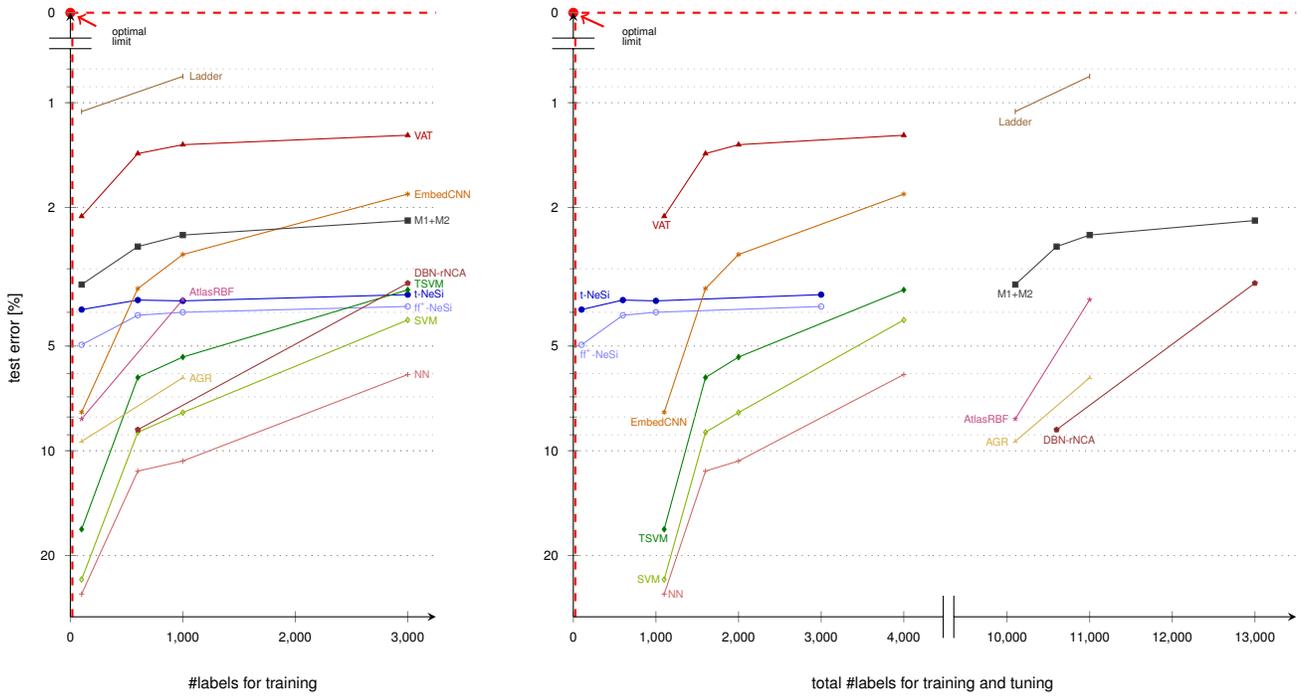

When comparing to state-of-the-art algorithms, we find that with decreasing numbers of labels the test error increases much slower for t-NeSi than for all competitors (see \cref{fig:MNIST-sota-Bar,fig:MNIST-sota-Scatter}).
Considering only training labels, NeSi networks are able to maintain low test errors even in the limit case of a single label per class (see \cref{tab:MNIST-results}), where no other model showed to operate so far.
Only very recent contributions come close to this here investigated limit case:
An ensemble of 10 improved generative adversarial networks (GANs) \cite{SalimansEtAl2016} showed state-of-the-art performance in the 100~training labels setting with a test error of $(0.86 \pm 0.56)\%$.
However, with only 20~training labels available, the test error already increased more than tenfold to $(11.34 \pm 4.45)\%$, where t-NeSi achieves $(7.22 \pm 0.53)\%$ with only 10~training labels.

Furthermore, all free parameters of the NeSi networks were optimized using only 10~labels per class in total, i.e., by using no more labels than available during training in the settings compared in \cref{fig:MNIST-sota-Bar,fig:MNIST-sota-Scatter}.
When considering the total amount of labels, used for the complete tuning and training procedure, we find that all competing models operate in settings of ten- to hundredfolds more labeled data as the NeSi networks (\cref{fig:MNIST-sota-Scatter}, right-hand side).
Although these additional labels were only used for parameter optimization and not training itself, it remains unclear how much their performance would degrade in lesser optimized parameter settings as overfitting effects can become drastic with higher model complexity and smaller validation sets.

%----------------------------------------------------------------
\subsection{NIST Special Database 19}
\begin{figure*}[!b]
    \centering
    \hfill
    \includegraphics[width=0.5\textwidth]{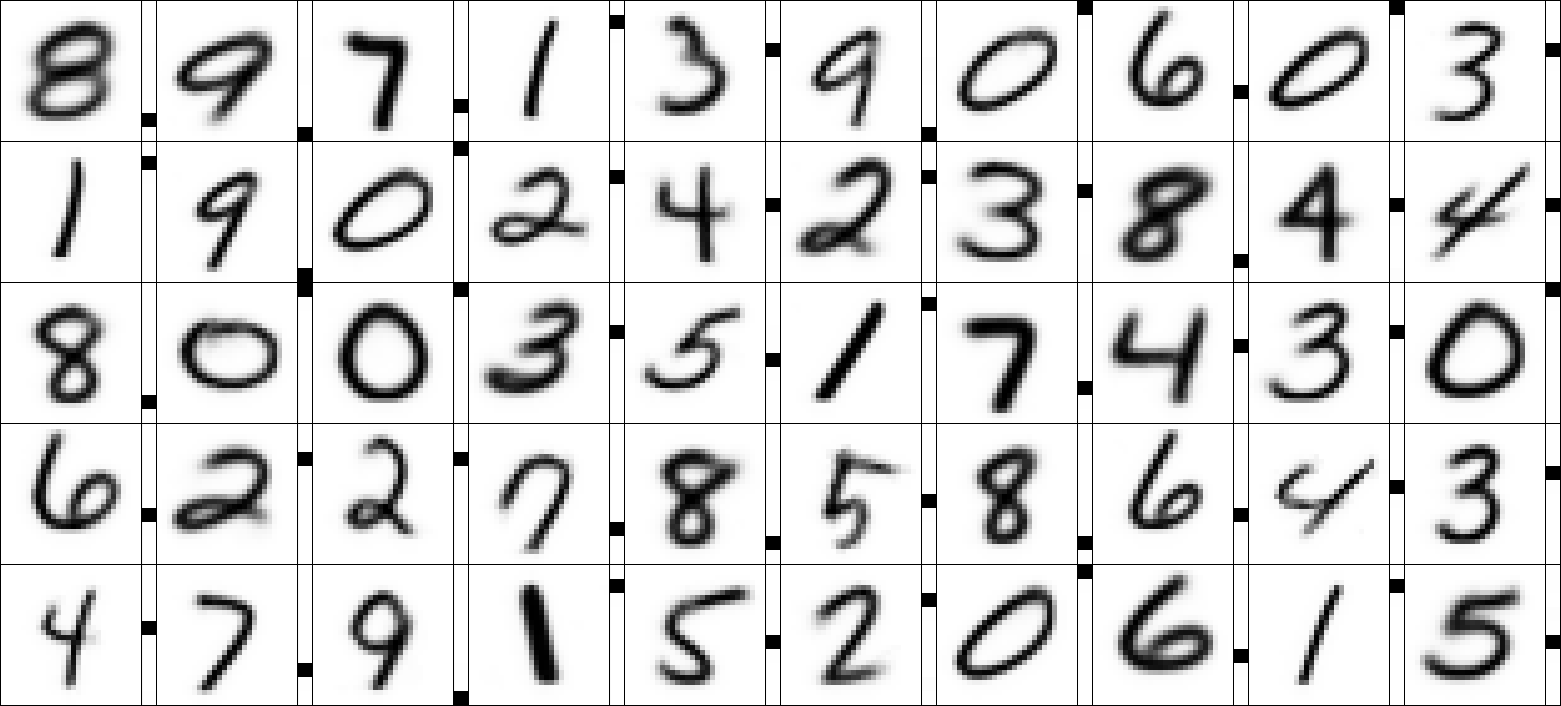}
    \hfill
    \includegraphics[width=0.463\textwidth]{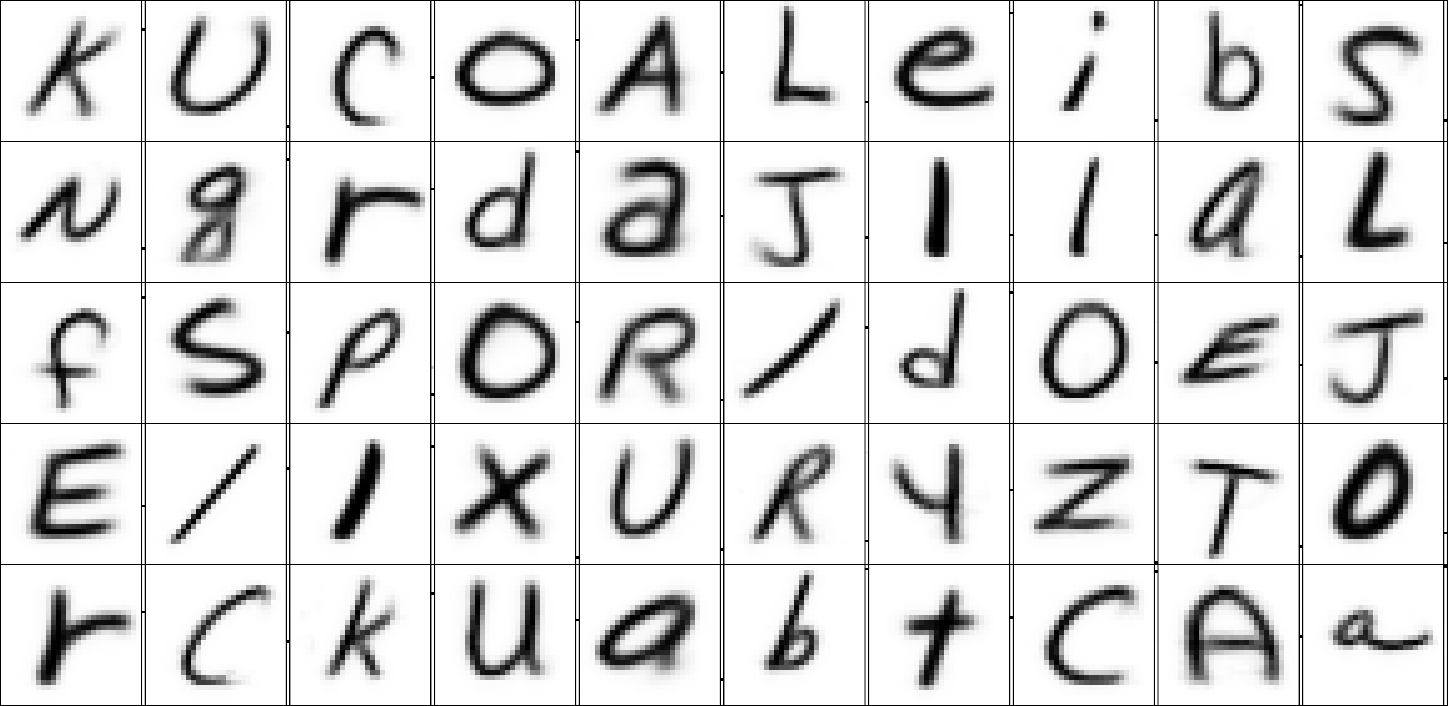}
    \hfill
    \caption{Visualization of a random subset of learned weights by the t-NeSi algorithm trained on NIST SD19 with only a single label per class on the task of digit (left-hand side) and character recognition (right-hand side). Shown are for both cases \num{50} of \num{10000} learned weights $\W[\subc,][:]$ of the first hidden layer as square matrices with weights $\R[:,][\subc]$ of the second hidden layer as columns next to them, indicating the learned class(es) of the individual fields (`0',\;\!\dots\;\!\!,`9' for digit data and `a',\;\!\dots\;\!\!,`z',`A',\;\!\dots\;\!\!,`Z' for characters).}
  	\label{fig:NIST-weights}
\end{figure*}

\begin{table*}[!tb]
  \centering
  {
\fontsize{8}{10} \sffamily \sansmath
\renewcommand{\arraystretch}{1.3}
\newcommand{\midspace}{\hphantom{abc}}
\setlength{\tabcolsep}{9pt}
\scriptsize
\begin{tabular}{@{}l r@{}c@{}l r@{}c@{}l r@{}c@{}l r@{}c@{}l r@{}c@{}l r@{}c@{}l r@{}c@{}l}
%\bottomrule
\toprule
\#labels/class & \multicolumn{3}{c}{1} & \multicolumn{3}{c}{10} & \multicolumn{3}{c}{60} &  \multicolumn{3}{c}{100} & \multicolumn{3}{c}{300} & \multicolumn{3}{c}{fully labeled}\\
\midrule
\noalign{\vskip 2pt} 
\multicolumn{6}{@{}l}{digits (10 classes)}\\
\#labels total & \multicolumn{3}{c}{10} & \multicolumn{3}{c}{100} & \multicolumn{3}{c}{600} & \multicolumn{3}{c}{1000} & \multicolumn{3}{c}{3000} & \multicolumn{3}{c}{344\,307} \\
\midrule
ff$^+$-NeSi \cite{ForsterEtAl2016} & 7.56 & $\,\pm\,$ & 1.79 & 6.20 & $\,\pm\,$ & 0.16 & 6.02 & $\,\pm\,$ & 0.08 & 6.02 & $\,\pm\,$ & 0.12 & 5.70 & $\,\pm\,$ & 0.03 & 5.11 & $\,\pm\,$ & 0.01 \\
t-NeSi & \bfseries  5.71 & $\,\pm\,$ & 0.42 & \bfseries 5.26 & $\,\pm\,$ & 0.23 & \bfseries 4.84 & $\,\pm\,$ & 0.02 & \bfseries 4.86 & $\,\pm\,$ & 0.03 & \bfseries 4.84 & $\,\pm\,$ & 0.02 & 4.50 & $\,\pm\,$ & 0.01 \\
35c-MCDNN\cite{CiresanEtAl2012} &&&&&&&&&&&&&&&& \multicolumn{3}{c}{\bfseries 0.77}\\
\midrule
\noalign{\vskip 2pt} 
\multicolumn{6}{@{}l}{letters (52 classes)}\\
    \#labels total& \multicolumn{3}{c}{52} & \multicolumn{3}{c}{520} & \multicolumn{3}{c}{3120} &  \multicolumn{3}{c}{5200} & \multicolumn{3}{c}{15\,600} & \multicolumn{3}{c}{387\,361}\\
	\midrule
			ff$^+$-NeSi \cite{ForsterEtAl2016} & 55.70 & $\,\pm\,$ & 0.62 & 46.22 & $\,\pm\,$ & 0.43 & 44.24 & $\,\pm\,$ & 0.23 & 43.69 & $\,\pm\,$ & 0.21 & 42.96 & $\,\pm\,$ & 0.28 & 34.66 & $\,\pm\,$ & 0.05 \\
   t-NeSi & \bfseries 52.14 & $\,\pm\,$ & 1.07 & \bfseries 45.62 & $\,\pm\,$ & 0.43 & \bfseries 41.87 & $\,\pm\,$ & 0.32 & \bfseries 41.75 & $\,\pm\,$ & 0.36 & \bfseries 41.13 & $\,\pm\,$ & 0.30 & 33.34 & $\,\pm\,$ & 0.04 \\
    35c-MCDNN\cite{CiresanEtAl2012} & & & & & & & & & & & & & & & & \multicolumn{3}{c}{\bfseries 21.01} \\
    \bottomrule
\end{tabular}
\setlength{\tabcolsep}{6pt}
}

  \vspace{8pt}
  \caption{Test error on NIST SD19 on the task of digit and letter recognition for different total amounts of labeled data. The results for NeSi are permutation invariant and given as the mean and standard error (SEM) over 10 independent repetitions, with randomly drawn, class-balanced labels.}
  \label{tab:NIST-results}
  \vspace{-10pt}
\end{table*}

The semi-supervised setting is especially interesting for practical applications where an abundance of unlabeled data is easily available but acquisition of accompanying labels requires vast amounts of additional human effort.
Algorithms in this domain should therefore be able to scale nicely with huge amounts of available unlabeled data.

To show the scalability properties of the truncated approach on hierarchical mixtures, we investigate the NIST Special Database 19 \cite{Grother1995}, which comprises \num{344307} training and \num{58646} test samples of handwritten digits ($\classK = 10$~classes) and \num{387361} training and \num{23941} test samples of handwritten characters ($\classK = 52$~classes).
We investigate semi-supervised settings of 1, 10, 60, 100 and 300 labels per class as well as the fully supervised setting on both tasks of digit and case-sensitive character recognition.
All results for t-NeSi are given as permutation invariant mean error rates on the respective blind test set over 10~independent runs with a new set of randomly picked, class-balanced training labels in each run.

\subsubsection{Preprocessing and Parameter Tuning}
We use the same data preprocessing as in \cite{ForsterEtAl2016}, which rescales, inverts and mass-centers the original $128 \times 128$ binary data to $28\times 28$ grayscale images, similar to MNIST (however without preserving relative scale).
Because of the high similarity, the free parameters for t-NeSi are kept the same as for MNIST, only scaling the learning rate of the top layer $\eR$ by a factor of five for faster convergence with the approximately fivefold as many unlabeled training samples.
The free parameters are then given as: $\normA=900$, $\subC=\num{10000}$, $\subC'=15$, $\eW=0.2 \times \subC/\inpN$, $\eR=1.0 \times \classK/\inpN$ and $\vartheta_{\BvSB}=0.6$, where $\classK = 10$ for digit classification and $\classK = 52$ for character classification.
As for MNIST, we again allow for 500 iterations over the respective training sets for sufficiently converged weights.
\Cref{fig:NIST-weights} shows an exemplary visualization of the learned weights.

\subsubsection{Results}
As for MNIST, we see significant improvements with truncated posteriors in all settings.
The strongest effect can again be observed in the settings of very few labels for the 10-class case of digit data (for which the free parameters were optimized on MNIST).
These results set a new baseline for permutation invariant NIST SD19 digit data in all here investigated semi-supervised settings and for letters data at \num{60}~labels per class and below (with r$^+$-NeSi \cite{ForsterEtAl2016} still being better in semi-supervised settings when more labels were available).
Notably, with additional unlabeled data the problem of class confusion in the limit case of a single label per class reduced even further with the truncated network and did not appear once in the here investigated 10~runs per setting.

For comparison, we show results of the state-of-the-art fully supervised 35c-MCDNN \cite{CiresanEtAl2012}, a committee of 35 deep CNNs.
Results for the semi-supervised setting were however only available for the NeSi networks \cite{ForsterEtAl2016}.

\section{Discussion}
Truncated variational EM provides a mathematical tool to derive truncated neural activations for generative networks, which we here applied to Neural Simpletrons.
The modified learning rules maintain a very compact and monolithic form and lead to reduced computational complexity compared to standard Neural Simpletrons \cite{ForsterEtAl2016}.
Usually, reducing the complexity of learning algorithms does result in reduced accuracy of obtained solutions.
Also in our case, performance decrease could have been expected.
However, as shown for MNIST in \cref{sec:NumericalExperiments}, we observe for truncated approaches ($\subC'=15$) consistently {\em higher} likelihood values for all EM iterations (see \cref{fig:MNIST-likelihood}).
Furthermore, these values are reached in fewer EM iterations than without truncation, which means an additional reduction of required computational cost to train Neural Simpletrons on a given data set.
This very beneficial scenario to simultaneously reduce computational cost and improve performance has been observed for truncated distributions previously: for ternary sparse coding \cite{ExarchakisEtAl2012} higher likelihood values were observed with truncation than without, and for binary and spike-and-slab sparse coding \cite{SheltonEtAl2011,SheikhLucke2016} likelihoods were observed to converge faster when using truncated distributions.
Also for mixture models \cite{DaiLucke2014,HughesSudderth2016}, truncated approximations have been observed to be beneficial for complexity and functionality and were explicitly reported to increase the likelihood faster in a currently ongoing study \cite{HughesSudderth2016}.
Like in our investigation, in all mentioned studies above, intermediate values for truncation parameters were found to perform best.

In this work, we, for the first time, applied truncated approximations to a minimalistically deep neural network and demonstrated improved semi-supervised classification performances.
Improved performance with reduced complexity is notably not unheard of for neural networks: dropout \cite{SrivastavaEtAl2014} is a popular approach showing similar features.
However, unlike dropout, which randomly discards usually about half of neural weights for updates to reduce co-adaptation of units, truncation systematically selects relatively few (and similar) weights for updates on a given data point.
Still, both approaches first considerably reduce the network's size before updating neural weights, which represents a remarkable similarity.
Notably, the here applied systematical truncation could still be combined with the random unit selection in dropout.

Based on the intuition for mixture models, by considering our numerical results, and based on explanations offered by sparse coding studies \cite{ExarchakisEtAl2012,SheltonEtAl2011,SheikhLucke2016}, truncated distributions seem to help avoiding local optima by destabilizing those locally optimal solutions that correspond to irrelevant solutions.
For sparse coding models \cite{ExarchakisEtAl2012,SheikhLucke2016}, such optima corresponded to relatively dense solutions which did not fit well to the investigated sparse data generation processes. 
For the here investigated mixture model, solutions corresponding to many strongly overlapping clusters are presumably discouraged.
At least for the data studied here, such distributions also do not seem to be beneficial for classification, which may explain the better performance with truncation.

In semi-supervised classification experiments, we saw strong beneficial effects in test error rates and variances, especially in the settings of very few labels.
With purely unsupervised learning in the first hidden layer, NeSi networks showed to be still applicable even in the limit case of only a single label per class during training, where truncated learning lead to relative improvements of up to 33\% in test error.
Maintaining low error rates without many labeled data points remains a huge challenge even for state-of-the-art deep networks, see, e.g., \cite{SalimansEtAl2016}.
This showed to be especially true, when tuning of free parameters is considered as additional optimization loop, where generally more additional labels are needed with higher model complexity (see \cref{fig:MNIST-sota-Bar,fig:MNIST-sota-Scatter}, right-hand side).

Scalability of the approach was investigated on the NIST SD19 database, which allows for training with more than fivefold as many (unlabeled) data points per class.
This data set shows to be more challenging than MNIST even when only considering digit data (as can be seen, e.g., by comparing state-of-the-art fully supervised performances of 35c-MCDNN \cite{CiresanEtAl2012}).
However, in the semi-supervised setting the additional unlabeled data lead to better defined templates (cluster centers) than for MNIST, which resulted in far slower degrading test errors with decreasing amounts of labels down to the very limit of a single label per class, even surpassing the MNIST results in that setting.
This shows, that the truncated NeSi network is very well suited for and greatly benefits from applications to data sets with very large amounts of unlabeled data.
We also investigated the much more challenging case-sensitive character classification task:
The number of classes here increases by a fivefold compared to digit classification, while the number of data points per class shrinks by approximately the same factor.
Also the overlap between classes increases because of higher similarities among the class means. Still, even without further parameter tuning, significant improvements could be observed for truncated NeSi also here, however less pronounced than for digit data.

In conclusion, our work is the first to demonstrate the advantages of truncated approximations in a neural network, where it further reduces complexity and improves on the already competitive performance
of NeSi networks for few labels.

% use section* for acknowledgment
\section*{Acknowledgment}
We acknowledge funding by DFG grant LU 1196/5-1 and by the Cluster of Excellence EXC 1077/1.

% trigger a \newpage just before the given reference
% number - used to balance the columns on the last page
% adjust value as needed - may need to be readjusted if
% the document is modified later
%\IEEEtriggeratref{8}
% The "triggered" command can be changed if desired:
%\IEEEtriggercmd{\enlargethispage{-5in}}

% references section

% can use a bibliography generated by BibTeX as a .bbl file
% BibTeX documentation can be easily obtained at:
% http://mirror.ctan.org/biblio/bibtex/contrib/doc/
% The IEEEtran BibTeX style support page is at:
% http://www.michaelshell.org/tex/ieeetran/bibtex/
%\bibliographystyle{IEEEtran}
% argument is your BibTeX string definitions and bibliography database(s)
%\bibliography{IEEEabrv,../bib/paper}
%
% <OR> manually copy in the resultant .bbl file
% set second argument of \begin to the number of references
% (used to reserve space for the reference number labels box)
{
\bibliographystyle{IEEEtran}
\bibliography{IEEEabrv,bibliography}
%\bibliography{IEEEabrv,bibliography}
}

% that's all folks
\end{document}